%% file: main.tex
\documentclass[10pt,twocolumn,letterpaper]{article}

\usepackage{iccv}
\usepackage{times}
\usepackage{epsfig}
\usepackage{graphicx}
\usepackage{amsmath}
\usepackage{amssymb}

\usepackage{tabularx}	
\usepackage{symbols}	
\usepackage[table,dvipsnames]{xcolor}
\usepackage{xspace}	
\usepackage{capt-of,etoolbox}	
\usepackage{footnote}
\usepackage{multirow}
\usepackage{booktabs}
\usepackage{colortbl}
\usepackage{wrapfig}
\usepackage{pifont}
\usepackage[export]{adjustbox}
\usepackage{multibib}
\newcites{supp}{{\large References}}
\usepackage{afterpage}
\usepackage[accsupp]{axessibility}
\usepackage{ogonek}

\usepackage[pagebackref=true,breaklinks=true,letterpaper=true,colorlinks,bookmarks=false]{hyperref}

\iccvfinalcopy %

\def\iccvPaperID{6640} %

\ificcvfinal\fi

\definecolor{ColorPNav}{HTML}{F5F5F5}
\definecolor{ColorFurnMove}{HTML}{FFF8D7}
\definecolor{ColorFootball}{HTML}{EFFFE8}

\newcommand{\unnat}[1]{{\color{purple}#1}}
\newcommand{\camera}[1]{{\color{black}#1}}
\newcommand{\as}[1]{{\color{orange}(Alex: #1)}}

\newcommand{\habitat}{\textsc{AIHabitat}\xspace}

\begin{document}

\title{\textsc{\gtop}: Training Embodied Agents with Minimal Supervision
}

\author{
Unnat Jain$^{1}$
\quad Iou-Jen Liu$^{1}$
\quad Svetlana Lazebnik$^{1}$
\quad Aniruddha Kembhavi$^{2}$\\
\quad Luca Weihs$^{2}$\thanks{denotes equal mentoring by LW and AS}
\quad Alexander Schwing$^{1}$\footnotemark[1]\\
\normalsize{$^{1}$University of Illinois at Urbana-Champaign\quad $^{2}$PRIOR @ Allen Institute for AI}\\
\normalsize{\url{https://unnat.github.io/gridtopix/}}
}

\maketitle

\input{sections/00_abstract}
\input{sections/01_introduction}

\input{sections/04_GridToPix}

\input{sections/05_experiments}

\input{sections/02_related}

\input{sections/06_conclusion}

{\small
\bibliographystyle{ieee_fullname}
\bibliography{unnat,supp}
}

\appendix
\input{sections/07_supplementary}
\end{document}

%% file: sections/00_abstract.tex
\begin{abstract}
While deep reinforcement learning (RL) promises freedom from  hand-labeled data,   great successes, especially for  Embodied AI, require significant work to create supervision via carefully shaped rewards.
Indeed, without shaped rewards, \ie, with only  terminal rewards, present-day Embodied AI results degrade significantly across Embodied AI problems from single-agent Habitat-based PointGoal Navigation (SPL drops from 55 to 0) and two-agent AI2-THOR-based Furniture Moving (success drops from 58\% to 1\%) to three-agent Google Football-based \footballtask (game score drops from 0.6 to 0.1). 
As training from shaped rewards doesn't scale to more realistic tasks, the community needs to improve the success of training with terminal rewards.  
For this we propose \gtop: 1) train agents with terminal rewards in gridworlds that generically mirror Embodied AI environments, \ie, they are independent of the task; 2) distill the learned policy into agents that reside in complex visual worlds. Despite learning from only terminal rewards with identical models and RL algorithms, \gtop significantly improves results across  tasks: from PointGoal Navigation (SPL improves from 0 to 64) and Furniture Moving (success improves from 1\% to 25\%) to football gameplay (game score improves from 0.1 to 0.6). \gtop even helps to improve the results of shaped reward training. %

\end{abstract}

%% file: sections/01_introduction.tex
\vspace{-2mm}
\section{Introduction} \label{sec:intro}
\begin{figure}[t]
    \centering
    \includegraphics[trim=10 5 10 3,clip,width=\linewidth]{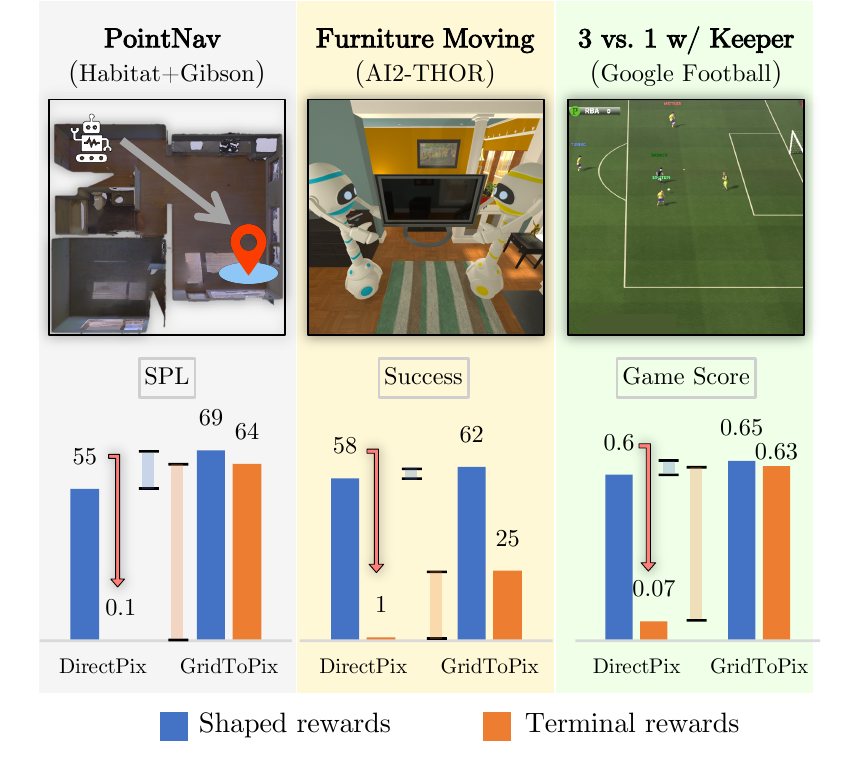}
    \vspace{-2mm}
    \caption{\textbf{Shaped \vs terminal rewards.} Across three tasks, embodied AI agents fail to learn from terminal rewards using standard RL methods despite achieving high performance when given carefully shaped rewards. When using our \gtop methodology, the embodied AI agents successfully learn from terminal rewards and  sometimes even outperform DirectPix trained agents receiving shaped rewards.}
    \vspace{-2mm}
    \label{fig:teaserNew}
\end{figure}

\begin{figure*}
    \centering
    \vspace{-1mm}
    \includegraphics[width=0.9\textwidth]{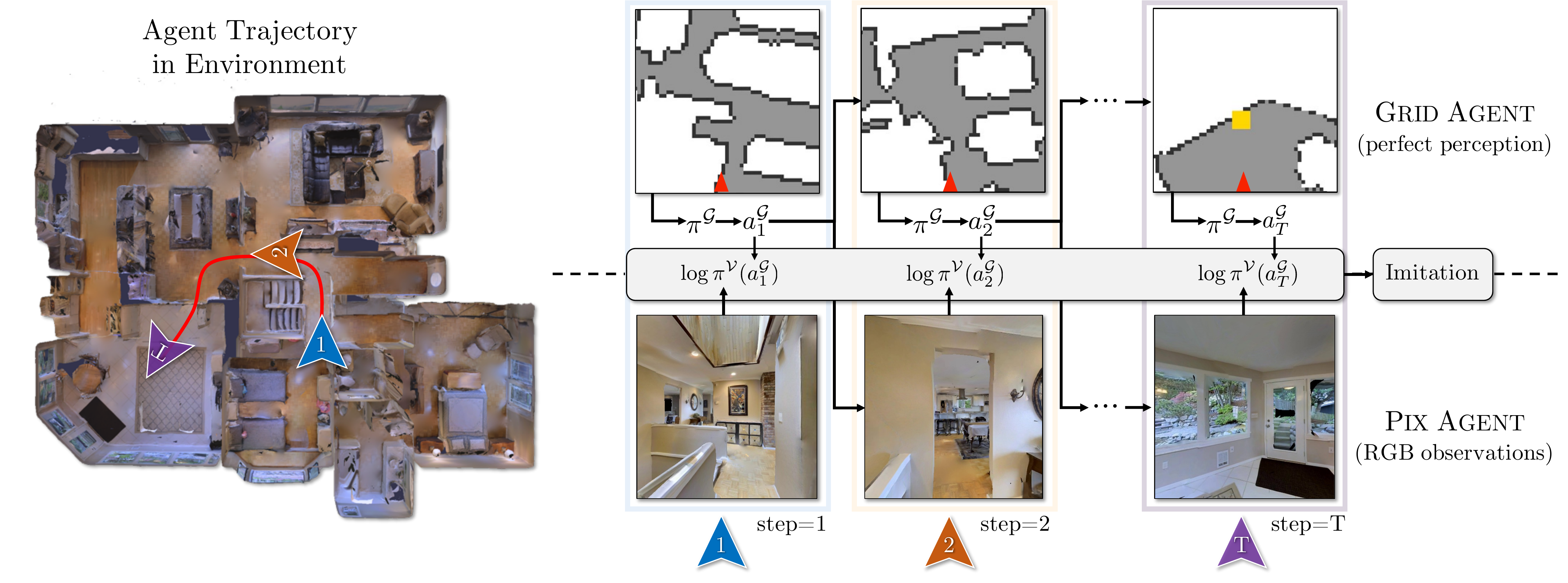}
    \vspace{-1mm}
    \caption{\textbf{\gtop methodology to transfer knowledge from gridworld to visual agents.}
    We propose to train agents in gridworld environments, where perception is simplified and simulation is fast, mirroring visual environments used in Embodied AI research. Learned gridworld policies can then be distilled to visual agents via imitation learning. See~\secref{sec:super-vis-agents-via-grid} for detailed notation.}\label{fig:teaser}
\end{figure*}
The Embodied AI research community has developed a host of capable simulated environments focused on the challenges of navigation~\cite{habitat19iccv,xia2018gibson}, interaction~\cite{ai2thor,igibson,Gan2020ThreeDWorldAP}, manipulation~\cite{SAPIEN,RLBench,MetaWorld}, and simulation-to-real transfer~\cite{robothor,pyrobot2019,Sim2RealHabitat}. Fast progress has been made within these environments over the past few years, particularly in navigation heavy tasks such as PointGoal navigation~\cite{habitat19iccv,wijmans2019dd}.

While early applications of deep RL to Embodied AI set down the more challenging path of learning from terminal rewards\footnote{\Ie, reward structures in which the only \emph{goal dependent reward} is given at the end of an episode. Goal independent rewards, \eg, a time-step penalty, can be given at every step.}~\cite{ZhuARXIV2016}, the pursuit of increasingly capable agents has steered us towards employing large amounts of human supervision~\cite{shridhar2020alfred,anderson2018vision,thomason2020vision,hahn2020you,ku2020room}, reward shaping~\cite{wijmans2019dd,wang2019reinforced}, and task-specific architectures~\cite{Chaplot2020Explore,ramakrishnan2020occant}. %
The multitude of rewards and auxiliary hyperparameters that are manually tuned when training Embodied AI agents today is reminiscent of %
careful feature engineering~\cite{sift,hog,Lazebnik2006BeyondBO} in computer vision years ago.

While these design choices have got our field (and agents) off the ground, it is hard to believe that this methodology will scale as  tasks increase in complexity and physical realism. Instead, we must borrow from the notable successes of reinforcement learning in games~\cite{Silver2017MasteringTG,Silver2018AGR}, where sophisticated agents are  trained with  minimal supervision in the form of terminal rewards.

As a first step towards this goal, we empirically analyze the ability of modern tried-and-true Embodied AI algorithms to learn high-quality policies from only  terminal rewards in visual environments (DirectPix). We consider a variety of challenging tasks in three diverse simulators -- single-agent PointGoal Navigation in Habitat~\cite{habitat19iccv} (SPL drops from 55 to 0), two-agent Furniture Moving in AI2-THOR~\cite{ai2thor} (success drops from 58\% to 1\%), and three-agent \footballtask in the Google Research Football Environment~\cite{GoogleResearchFootball} (game score drops from 0.6 to 0.1). Given only  terminal rewards, we find that performance of these modern methods degrades drastically, as summarized in Fig.~\ref{fig:teaserNew}. Often, no meaningful policy is learned despite training for millions of steps. These results are eye-opening and a reminder that we are still ways off in our pursuit of embodied agents that can %
learn skills with minimal supervision, \ie, supervision in the form of only terminal rewards.

If present-day RL algorithms have been successful in other domains (particularly ones requiring less visual processing~\cite{Silver2017MasteringTG,Silver2018AGR,babyai_iclr19}), why do they struggle for Embodied AI? Our hypothesis: this struggle is due to the need for embodied agents to learn to plan and perceive simultaneously. This introduces a `non-stationarity' into  learning -- the planning module must continuously adapt to changes in perceptual reasoning. While non-stationarity in learning has also been a problem in past challenges conquered by RL methods, Embodied AI tasks exacerbate this problem due to the presence of rich and diverse visual observations, long-horizon planning, a need for collaboration, and a requirement to generalize to  unseen test worlds.

Driven by this hypothesis, we study \gtop, a 
training routine
for embodied agents that decouples the joint goal of planning from visual input into two manageable pieces. Specifically, using general-purpose \textit{gridworld} environments which generically mirror the embodied environments of interest, we first train a gridworld agent for the desired task.
Within a gridworld, an agent has perfect visual capabilities, allowing learning algorithms to focus on long-horizon planning %
given only terminal rewards. Next, this gridworld agent supervises an agent operating \camera{\textit{solely}} on complex visual observations \camera{(no gridworld is needed at test time). 
Importantly, our conceptualization of `gridworld' isn't restricted to a top-down occupancy map. As we detail in \secref{sec:thor-habitat-gridworlds}, gridworlds are perfect-perception environments, \ie, \textit{semantics are explicitly available} to the agent.} 

Across tasks, as summarized in~\figref{fig:teaserNew}, we observe that \gtop significantly outperforms directly comparable\footnote{\Ie, using identical model architectures.} methods when training visual agents using terminal rewards: for PointGoal Navigation the SPL metric improves from 0 to 64; for Furniture Moving success improves from 1\% to 25\%; for \footballtask the game score improves from 0.1 to 0.6.  
Moreover, \gtop %
even improves
benchmarks when trained with carefully shaped rewards. 
This finding is analogous to the progress made by weakly supervised computer vision approaches which inch towards benchmarks set by fully supervised methods.

%% file: sections/04_GridToPix.tex
\section{Why Learn From Terminal Rewards?}
Human supervision, hand-written and rule-based expert teachers, shaped rewards, and custom architectures are common when developing increasingly capable embodied agents. Several questions naturally arise when developing %
agents with only minimal supervision.

\noindent\textit{As rule-based optimal experts are frequently available in today's simulated environments and tasks, why not use them for dense supervision at every step?} Admittedly, many present-day embodied tasks are navigational %
which permits computing optimal actions easily using shortest path algorithms. The community is, however, quickly moving to more intricate and physics-based tasks, where rule-based experts are %
computationally expensive or infeasible.
Consider, for example, the games of Go, Hanabi, and Football, or the real world tasks of elderly assistance and disaster relief. It is extremely difficult to devise a heuristic expert, let alone an optimal one, for any of those examples. Creating heuristic experts in these settings is just as, if not more, labor intensive as the reward shaping we hoped to avoid. With these long term pursuits in mind, we think it is important  to study and develop methods that  learn with minimal supervision, including in tasks that are popular today. In this work, optimal actions can be easily computed for the popular Point Navigation task~\cite{habitat19iccv}. But for the other two tasks we consider -- \fmove~\cite{JainWeihs2020CordialSync} and \footballtask~\cite{GoogleResearchFootball} -- no experts are available in the literature and creating such an expert appears to be extremely difficult.

\noindent\textit{If not rule-based experts, how about collecting human annotations?} Deep models need the equivalent of years of expert supervision even for perfect-perception tasks~\cite{babyai_iclr19}.\footnote{BabyAI gridworld~\cite{babyai_iclr19} studies show that ${\sim} 21$M expert actions are needed to train an agent to complete an instruction following task with high-performance. If humans  produce 1 action per second,  246 days of labeling are required.} Embodied AI would undoubtedly require a similar (if not much larger) number of annotations. Collecting these human annotations is labor intensive and must be done for every new task and behavior we wish our agent to complete. This is extremely expensive and intractable for more complex behaviors. In contrast, terminal rewards are easy to provide and allow  years of simulated self-play (tractable in wall-clock time). This goal aligns closely with the pursuits of the AI community -- pre-training with self-supervision and then learning new skills with minimal human help.

\noindent\textit{Why should we expect that computing terminal rewards is easier than computing shaped rewards?} While we can't ``prove'' that terminal rewards are easier to compute than shaped rewards, we believe there is good evidence that this is, in general, the case. Shaped rewards are frequently designed so that they provide, at every step, feedback to the agent about its progress towards achieving the goal. Computing such rewards thus requires  to approximate a `distance,' in terms of agent actions, between an arbitrary state and the goal. It is not hard to construct tasks where computing such a distance is NP-complete (exponential-time with known algorithms). 
In comparison, verifying that an agent has reached a goal (and thus deserves its terminal reward) requires no such search, making it, in principle, an easier problem.\footnote{The entire class of NP-complete problems is one in which verifying solutions is easy (polynomial time) but in which finding these solutions is fundamentally harder (unless P$=$NP).}

\section{\gtop}
\label{sec:app}
We are interested in learning from only terminal rewards a high-performance policy $\mv$ which acts on realistic (\eg, visual) observations within an Embodied AI environment, hereafter called the visual environment. %
However, empirically and irrespective of the task and environment, we find joint learning of perception and planning from terminal rewards (DirectPix) to be extremely challenging, as summarized in Fig.~\ref{fig:teaserNew}.

In contrast, due to their perceptual simplicity, gridworlds are generally extremely fast to simulate, and learning is sample efficient as a gridworld policy $\mg$ needs to devote little effort to learn accurate perception. %
This enables  gridworld agents to rapidly learn high-quality policies $\mg$ from only terminal rewards.

This advantage of gridworlds is an opportunity to reduce labor-intensive reward shaping efforts: can we first learn policies from terminal rewards within gridworlds that replicate the dynamics of a visual environment and then efficiently transfer these policies to visual agents?

We refer to the approach which is designed to enable this transfer and which is sketched in Fig.~\ref{fig:teaser} as \gtop, and discuss it next. 
In particular, Sec.~\ref{sec:terminal-vs-shaped-rewards}  formalizes terminal and shaped rewards, Sec.~\ref{sec:thor-habitat-gridworlds} describes how we design gridworlds that replicate a visual environment's dynamics, and Sec.~\ref{sec:super-vis-agents-via-grid} details both how we train gridworld agents and how imitation learning can be used to transfer gridworld policies to visual agents.

\subsection{Terminal \vs Shaped Reward Structures}\label{sec:terminal-vs-shaped-rewards}
The reward structure used during training has a substantial impact on  sample efficiency, stability, and quality of the learned policy. Most prior work for the tasks considered in this paper used a shaped reward structure, \ie, the reward obtained by the agent at timestep $t$ can be decomposed into
\begin{align}
    r^{\text{shaped}}_t =
     r^{\text{success}}_t + r^{\text{progress}}_t + r^{\indep}_t \quad. \label{eq:rewards_dense}
\end{align}
Here, $r^{\text{success}}_t$ is a sparse terminal reward equalling 0 on all but the last timestep; for our considered tasks,
\begin{align}
    \hspace{-0.2cm} r^{\text{success}}_t \!\!=\!  \begin{cases}
     \!r_{\text{success}}\cdot(1 \!-\! \alpha\cdot \frac{t}{T}) & \hspace{-0.3cm}\text{if goal achieved,}\\
     \!0 & \hspace{-0.3cm}\text{otherwise,}
    \end{cases}\label{eq:rewards_terminal}
\end{align}
where $T\geq 1$ is the maximum allowed episode length and $r_{\text{success}}> 0$, $\alpha\in[0,1]$ are task-specific constants.\footnote{The inclusion of the $\alpha$ parameter in terminal rewards allows to encode a more nuanced measure of success. Note that when  $\alpha>0$, $r^{\text{success}}_t$ is larger when the agent completes the task in fewer steps.}
Further, $r^{\text{progress}}_t$ is a dense, goal-dependent reward, often equalling the change in distance to the goal. %
Finally, $r^{\indep}_t$ is a possibly dense but goal-independent reward, meant to encourage behaviors that are generally thought to be ``good'' regardless of the task, \eg, $r^{\indep}_t$ often includes a small negative step penalty which gently encourages the agent to complete its task quickly. 

As previously discussed, the goal-dependent shaped rewards $r^{\text{progress}}_t$ must be manually designed, can be computationally expensive, and generally require privileged access to the environment state. Because of this, we often aim to train  agents with the simpler reward structure
\begin{align}
    r^{\text{terminal}}_t =
     r^{\text{success}}_t + r^{\indep}_t \quad. \label{eq:rewards_simple}
\end{align}
In fact, for our experiments with terminal rewards, we let $r^{\indep}_t \equiv 0$  in all but the \fmove task where we keep prior work's inclusion of a step and failed-coordination penalty.

\subsection{Gridworld Mirrors of Visual Environments}
\label{sec:thor-habitat-gridworlds}

In order for policies obtained from gridworld training to be successfully transferable to visual agents, we need these gridworlds to be mirrors of their visual counterparts: a step in the gridworld should be translatable to a step in the visual world. 
We will now describe the three gridworlds used in our experiments, mirroring the \habitat, \thor, and Google Football environments. While these gridworlds work well for the considered tasks, they are also applicable for other tasks trained in these environments. We hope to encourage current and future Embodied AI environment developers to provide comprehensive general purpose gridworlds corresponding to their environments.

\noindent\textbf{\textsc{Grid-Habitat}.} 
We extend the existing functionality of the \habitat platform to permit, for the first time,  training of gridworld agents. Specifically, we generate an egocentric, top-down, observation with the agent positioned at the bottom facing towards the center (see Fig.~\ref{fig:teaser}). This top-down observation contains sensor information about occupancy (\ie, free space and walls) and goal location. This sensor information is sufficient for \pnav and can be easily enriched for other tasks (\eg, by adding semantic channels for ObjectNav). Our gridworld observation is different from the top-down visualization tool provided by \habitat which is allocentric and scaled differently for different scenes. We call this new gridworld \textsc{Grid-Habitat}. 

\noindent\textbf{\textsc{Grid-AI2-THOR}.} For the multi-agent {\fmove} task, we follow Jain~\etal~\cite{JainWeihs2020CordialSync} and conduct our study on an \thor-mirroring top-down gridworld.
Given the complexity of \fmove, we include information in separate channels of the top-down tensor with each channel corresponding to the output of a distinct sensor. These sensors specify whether a cell can be occupied by a furniture item, whether the cell is reachable by an agent, whether it was previously visited, the location of the furniture item, and the location of the other agent. We refer to this gridworld as \textsc{Grid-\thor}.

\noindent\textbf{\textsc{Grid-Google-Football}.} For the multi-agent Google football environment 
created by Kurach~\etal~\cite{GoogleResearchFootball}, we construct an observation that summarizes important game state information. Specifically, 
for each controlled player, the gridworld observation is a 1D vector which contains the location of all other players, the ball, and the opponent goal relative to the controlled player's location.

\camera{Importantly, as suggested in \secref{sec:intro}, our conceptualization of a gridworld goes beyond a spatial top-down occupancy map. While observations of \textsc{Grid-Habitat} capture occupancy and goal information, the observation of \textsc{Grid-AI2-THOR} is a tensor containing \textit{explicit semantics} of the environment. Moreover, as in \textsc{Grid-Google-Football}, gridworld observations need not be restricted to have spatial structure. Here, as is common in RL literature, gridworld agents receive a 1D vector capturing perfect-perception observations.}

\begin{figure}
    \centering
    \includegraphics[width=0.9\linewidth]{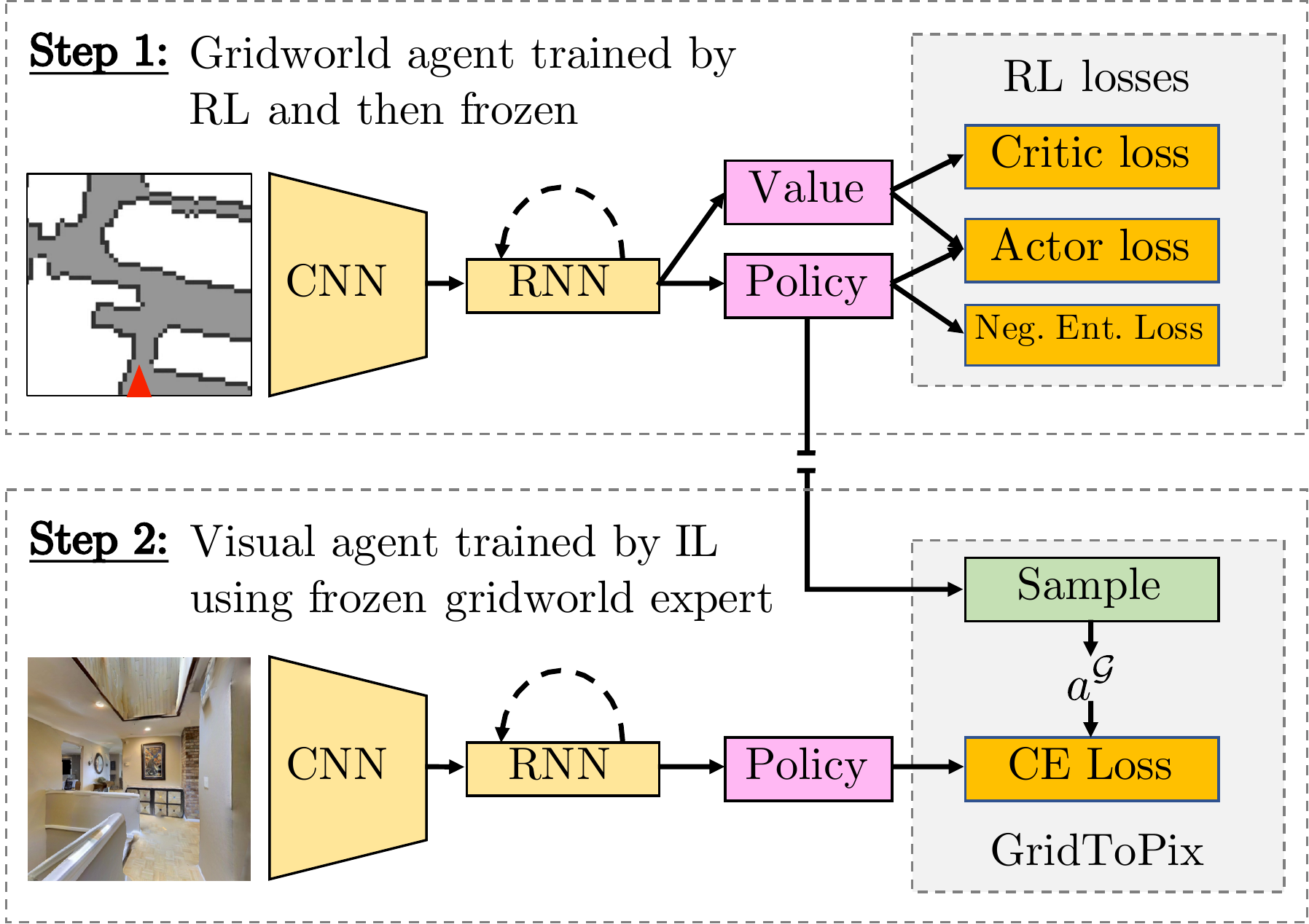}
    \vspace{-1mm}
    \caption{\textbf{Model overview.} Schematic of how top-down grid and visual observations are processed by their respective networks. The gridworld agent is trained via actor-critic losses, and subsequently supervises the visual agent via a cross-entropy loss.
    }
    \label{fig:overview}
\end{figure}

\subsection{Supervising Visual Agents via Gridworlds}\label{sec:super-vis-agents-via-grid}

We now describe how we train  gridworld agents and subsequently use imitation learning to transfer their learned policies to visual agents. An overview is given in Fig.~\ref{fig:overview}.

\noindent\textbf{Training in Gridworlds.}
As is standard in RL, we train gridworld agents to maximize the expected $\gamma$-discounted cumulative reward of their actions. Task specific details (\eg, algorithms and hyperparameters) are included in~\secref{sec:tasks}.
Whenever possible we follow protocols laid out by prior work.

\noindent\textbf{Distilling policies.}
How can we effectively train the parameters of a visual agent policy $\mv$ given a policy $\mg$ trained in a gridworld? 
To answer this question, recall that the visual agent $\av$ is interacting with a visual environment for which every step is translatable to a gridworld. The visual agent can hence be supervised by a gridworld agent $\ag$. 
The visual agent $\av$ might act using its own policy $\mv$ or adopt an \textit{exploration policy} $\mu$. This leads to on-policy and off-policy variants of imitation learning (IL). To illustrate this, let a rollout of the visual agent $\av$ and the  gridworld agent $\ag$  be $(a_1^{\av}, a_2^{\av},  \dots, a_t^{\av})$ and $(a_1^{\ag}, a_2^{\ag}, \dots, a_t^{\ag})$, respectively. To train the parameters of the visual policy $\mv$ via \gtop, we employ  the  cross-entropy loss. Formally, suppose that $O_\mu$ is a random variable corresponding to the observation seen by an agent when following policy $\mu$ and suppose that $H_\mu$ is a random variable encoding the history of all observations seen before obtaining observation $O_\mu$. The \gtop loss is then
\begin{equation}
    \mathcal{L}_{\text{\gtop}} = \mathbb{E}[\mathbb{E}_{a\sim\mg(\cdot \mid O_\mu, H_\mu)}[-\log\mv(a\mid O_\mu, H_\mu)]]. 
\end{equation}
The choice of exploratory policy $\mu$ leads to three variants, each widely adopted in IL %
tasks: %
\camera{ student forcing, teacher forcing, and annealed teacher forcing or DAgger (details in Appendix).}
In our experiments we primarily use DAgger.
Importantly, no matter the exploration policy used during training, at test time the visual agent's policy $\mv$ is deployed so that there is no access to gridworld policies at test time.

%% file: sections/05_experiments.tex
\section{Tasks, Models, and Evaluation}
\label{sec:tasks}
We evaluate \gtop using three tasks. We  chose these tasks as they (a) span both single-agent and multi-agent settings, (b) include tasks that do not have easily constructed rule-based experts (\fmove and \footballtask Football) as well as a task for which optimal actions can be computed from shortest path computations (\pnav), (c) provide the opportunity to test \gtop across a variety of different reward structures and model architectures, and (d) employ three different Embodied AI environments. 
\camera{Additionally, in the Appendix, we include experiments for a task where \textit{perfect reward shaping is intractable} -- Visual Predator-Prey (in OpenAI multi-particle environment~\cite{LoweNIPS2017,MordatchAAAI2018}).}
Across all experimental results, we present %
standard evaluation metrics after 10\% and 100\% of training has completed (to provide an understanding of sample efficiency and asymptotic results).

\subsection{PointGoal Navigation}
\label{sec:task-pnav}
PointGoal Navigation (\pnav) is a single agent navigation task specified for the \habitat\ simulator. An agent is spawned at a random location in the scene and must navigate to a target location specified by  coordinates relative to the agent's current location.  The relative goal position is available at every time step and the agent navigates by choosing one of four actions at every step.

\noindent\textbf{Shaped rewards.} Traditionally trained with shaped rewards~\cite{habitat19iccv,wijmans2019dd,ye2020auxiliary}, the change in shortest-geodesic-path distance to the goal is chosen as the goal-dependent progress reward $r^{\text{progress}}_t$ (see~\equref{eq:rewards_dense}). Computing it requires access to the full scene graph and a shortest-path planner. Here the success reward $r^{\text{success}}_t$ is chosen as in Eq.~\eqref{eq:rewards_terminal} with $r_{\text{success}}=10$ and $\alpha=0$. Finally, $r^{\indep}_t$ is a constant step penalty of $-0.01$.

\noindent\textbf{Terminal rewards.} Here we use the reward structure from \equref{eq:rewards_simple} with $r_t^\text{success}$ obtained from 
\equref{eq:rewards_terminal} with $r_{\text{success}}=10$ and $\alpha=0.9$. We set $r_t^{\indep} = 0$.

\noindent\textbf{Model architecture.} For a fair comparison, we use the standard CNN-GRU architecture~\cite{habitat19iccv,wijmans2019dd,chen2019audio,cartillier2020semantic,chang2020semantic,wani2020multion}. We adopt the official  implementation\footnote{github.com/facebookresearch/habitat-lab} and train with PPO~\cite{schulman2017proximal}, the de-facto standard RL algorithm for \pnav.

\noindent\textbf{Evaluation.} \pnav agents are primarily evaluated via Success weighted by Path Length (SPL)~\cite{anderson2018evaluation} and percentage of successful episodes (Success). We set a budget of 50 million environment steps\footnote{${\sim}2.5$ days of training using a \emph{g4dn.12xlarge} AWS instance configured with 4 NVIDIA T4 GPUs, 48 CPUs, and 192 GB memory.} and report results on the Gibson validation set of 14 unseen scenes with 994 episodes.

\begin{table*}[t]
\vspace{-0.2cm}
\centering
\resizebox{\textwidth}{!}{%
\begin{tabular}{l>{\columncolor{ColorPNav}}c>{\columncolor{ColorPNav}}c>{\columncolor{ColorPNav}}c>{\columncolor{ColorPNav}}c>{\columncolor{ColorFurnMove}}c>{\columncolor{ColorFurnMove}}c>{\columncolor{ColorFurnMove}}c>{\columncolor{ColorFurnMove}}c>{\columncolor{ColorFootball}}c>{\columncolor{ColorFootball}}c}

\toprule
\multicolumn{1}{c}{Tasks $\rightarrow$} & \multicolumn{4}{c}{\cellcolor{ColorPNav}\textbf{PointGoal Navigation}} & \multicolumn{4}{c}{\cellcolor{ColorFurnMove}\textbf{Furniture Moving}} & \multicolumn{2}{c}{\cellcolor{ColorFootball}\textbf{3 \vs 1 with Keeper}} \\
\multicolumn{1}{c}{} & \multicolumn{2}{c}{\cellcolor{ColorPNav}SPL} & \multicolumn{2}{c}{\cellcolor{ColorPNav}Success} & \multicolumn{2}{c}{\cellcolor{ColorFurnMove}MD-SPL} & \multicolumn{2}{c}{\cellcolor{ColorFurnMove}Success} & \multicolumn{2}{c}{\cellcolor{ColorFootball}Game Score} \\
Training routines $\downarrow$ & \emph{@10\%} & \emph{@100\%} & \emph{@10\%} & \emph{@100\%} & \emph{@10\%} & \emph{@100\%} & \emph{@10\%} & \emph{@100\%} & \emph{@10\%} & \emph{@100\%} \\
\midrule
\directrl & 0.0 & 0.1 & 0.0 & 0.2 & 0.0 & 0.0 & 0.8 & 0.8 & 0.03 & 0.07 \\
\gtop & \textbf{45.4} & \textbf{63.8} & \textbf{63.5} & \textbf{81.8} & 1.6 & \textbf{4.0} & 16.4 & \textbf{24.6} & 0.33 & \textbf{0.63} \\
\gtop $\rightarrow$ \directrl & 44.6 & 59.7 & 62.1 & 77.7 & \textbf{2.7} & 3.1 & \textbf{19.8} & 14.5 & \textbf{0.35} & 0.6 \\
\midrule
Gridworld expert (\textit{upper bound}) & \multicolumn{2}{c}{\cellcolor{ColorPNav}\textit{78.8}} & \multicolumn{2}{c}{\cellcolor{ColorPNav}\textit{94.2}} & \multicolumn{2}{c}{\cellcolor{ColorFurnMove}\textit{19.2}} & \multicolumn{2}{c}{\cellcolor{ColorFurnMove}\textit{56}} & \multicolumn{2}{c}{\cellcolor{ColorFootball}\textit{0.9}} \\
\bottomrule
\end{tabular}%
}
\vspace{0.03in}
\caption{\textbf{Quantitative results (terminal reward structure).} Metric values for the \pnav, \fmove, and \footballtask tasks reported on their respective evaluation sets (see~\secref{sec:tasks}). Across all the three tasks, the \gtop agents outperform their \Directrl counterparts. For ease of reading, SPL is scaled by 100 and success \% is reported. To quantify the efficiency of learning, metrics are reported after \textit{10\%} and \textit{100\%} of training has completed. The last row is separated to highlight that gridworld experts serve as a loose \textit{upper bound} for \gtop agent performance and should otherwise not be directly compared against.
}
\label{tab:terminal-metrics}
\end{table*}

\begin{figure*}[t]
\vspace{-0.3cm}
    \centering
    \begin{tabular}{ccc}
        \multicolumn{3}{c}{\includegraphics[width=0.7\textwidth]{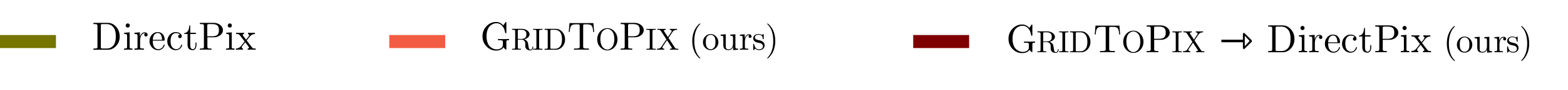}}\vspace{-2mm}\\
        \includegraphics[trim=5 5 10 10,clip,width=0.3\textwidth]{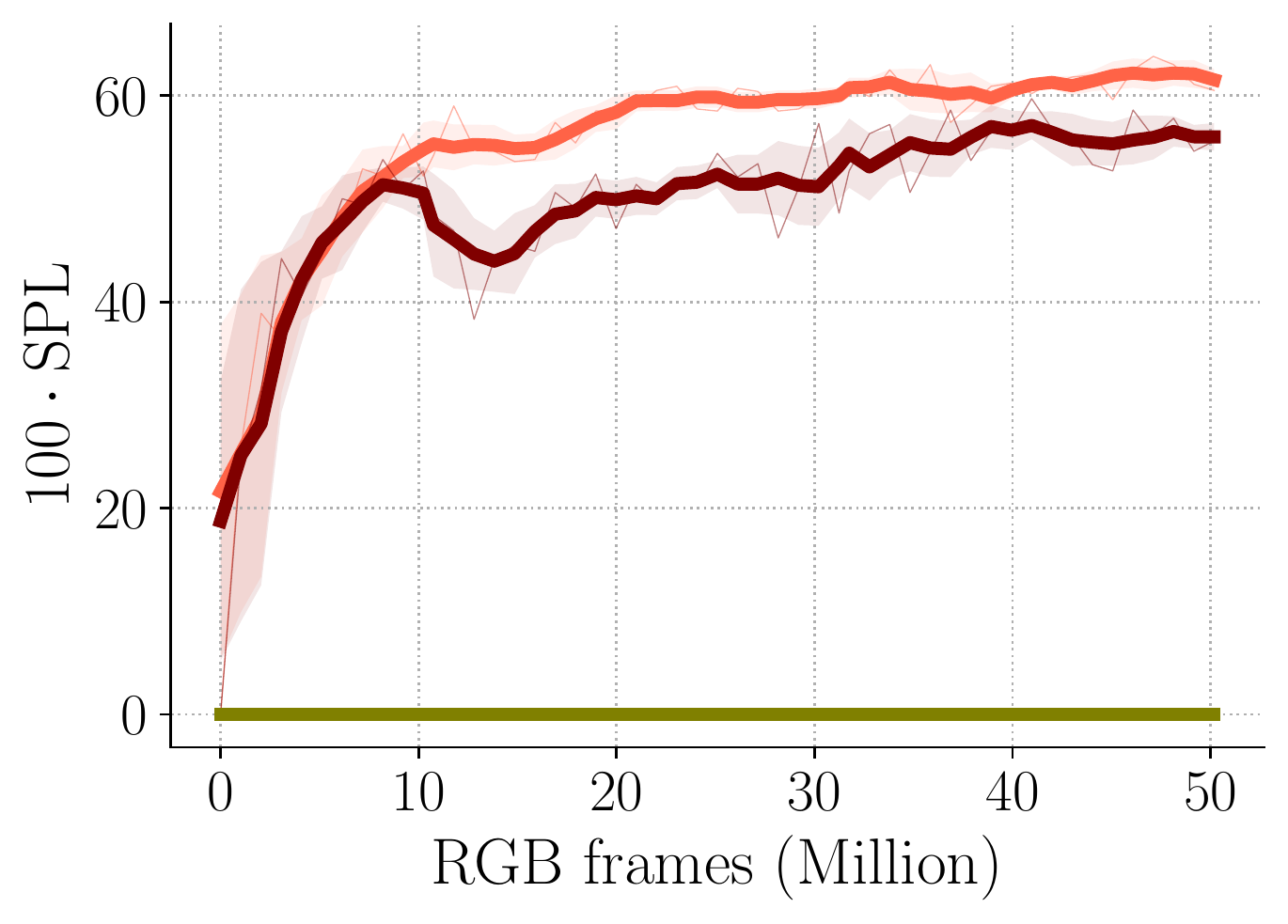} &
        \includegraphics[trim=5 5 10 10,clip,width=0.3\textwidth]{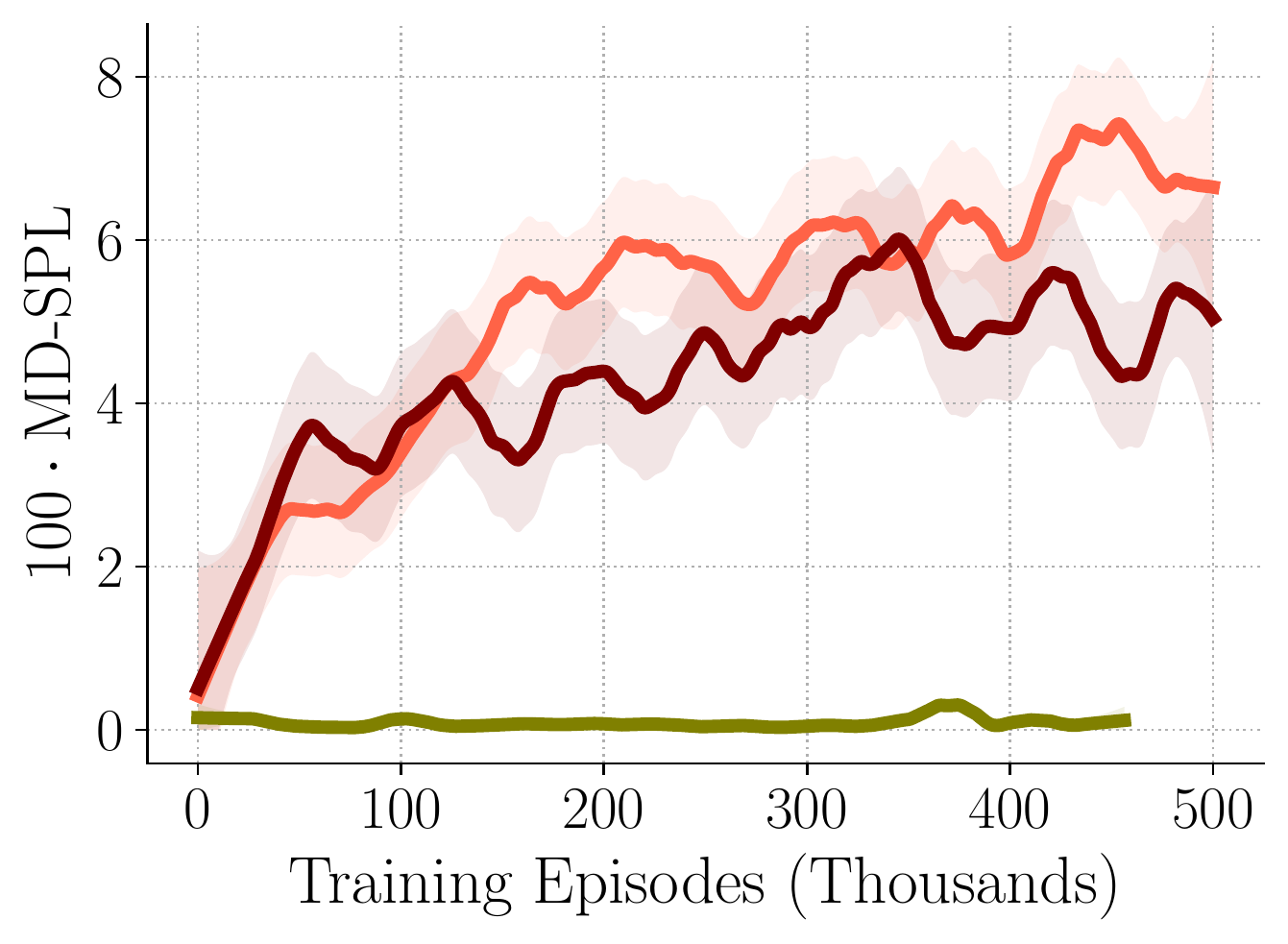} &
        \includegraphics[trim=5 5 10 10,clip,width=0.3\textwidth]{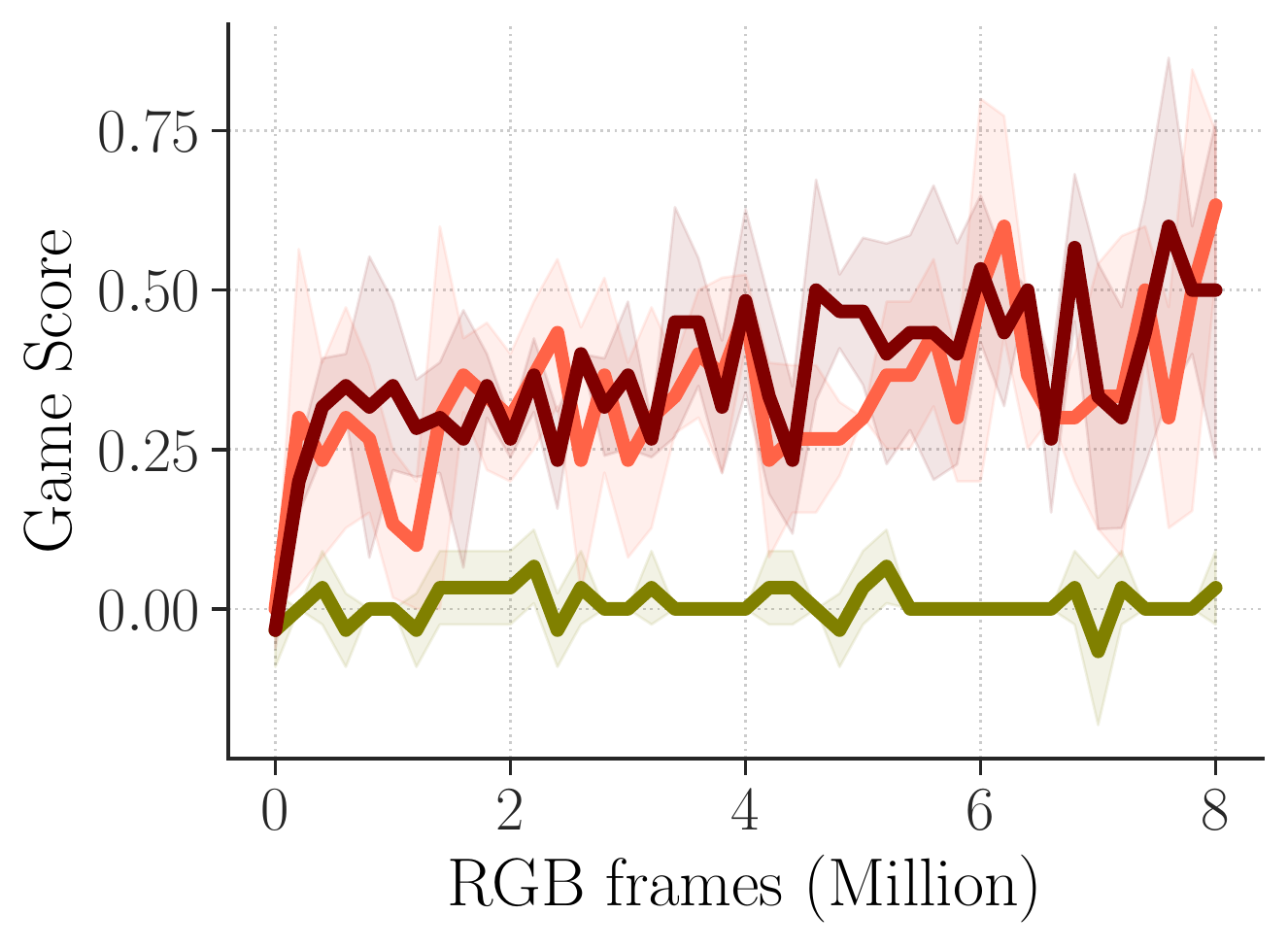}\\
        (a) PointGoal Navigation & (b) Furniture Moving & (c) Football -- \footballtask
    \end{tabular}
    \vspace{0.03in}
    \caption{\textbf{Learning curves on validation set (terminal reward structure).} Primary metrics are plotted \vs training steps/episodes, following the standard protocols for the respective task (see~\secref{sec:tasks}).
    (a) Thin lines mark checkpoints evenly spaced by $1$ Mn frames. Bold lines and shading mark the rolling mean (with a window size of $2$) and the corresponding standard deviation. (b) In line with~\cite{JainWeihs2020CordialSync}, we log information by running an online validation process and employ a local quadratic regression smoothing with 95\% confidence intervals. (c) The plot shows the average game score and standard deviation. Checkpoints are evenly spaced by $200$K frames.
    Despite considerable effort, \Directrl methods  fail to learn meaningful policies with terminal rewards.
    }
    \label{fig:terminal-plots}
    \vspace{-0.1in}
\end{figure*}

\subsection{Furniture Moving}
\label{sec:task-fmove}
\task is a challenging two-agent furniture moving task set within the \thor\ simulator~\cite{JainWeihs2020CordialSync}. Two agents collaborate to move an object through the scene and place it above a visually distinct target. Agents may communicate with other agents at each timestep using a low bandwidth communication channel. Each agent can choose from 13 actions. Specifically, in addition to the vanilla navigation, agents may move with the lifted object, move just the object, and rotate the object. Moreover, due to two agents, the joint action space contains 169 actions. 

\noindent\textbf{Shaped rewards.} Here, change in Manhattan distance to the goal\footnote{In \fmove, computing the shortest-geodesic-path is intractable as each location of the furniture corresponds to over 400k states~\cite{JainWeihs2020CordialSync}.} is used as the goal-dependent reward $r^{\text{progress}}_t$. Three goal-independent rewards are suggested in~\cite{JainWeihs2020CordialSync} to help learn the coordinated policy for this multi-agent system. In particular, $r^{\indep}_t$ includes a step penalty, a joint-pass (or do-nothing) penalty, and a failed action penalty. 

\noindent\textbf{Terminal rewards.} For a head-on comparison with~\cite{JainWeihs2020CordialSync} we simply drop $r^{\text{progress}}_t$. Like prior work~\cite{JainWeihs2020CordialSync}, the success reward $r_t^\text{success}$ is obtained from \equref{eq:rewards_terminal} using  $r_{\text{success}}=1$ and $\alpha=0$.
All \fmove results are based on this. Additional results with $r^{\indep}_t=0.01$ are included in the appendix.

\noindent\textbf{Model architecture.} We utilize SYNC, the best-performing  architecture from prior work~\cite{JainWeihs2020CordialSync}.
For fairness, we use the same RL algorithm (A3C~\cite{MnihEtAlPMLR2016}) as prior work.

\noindent\textbf{Evaluation.} \task agents are primarily evaluated via two metrics -- \% successful episodes (Success) and a Manhattan distance based SPL (MD-SPL). Additional metrics are reported in the appendix. Consistent with~\cite{JainWeihs2020CordialSync}, we train for 500k episodes for the visual agents.\footnote{2-4 days of training with a \emph{g4dn.12xlarge} AWS instance.} Metrics are reported on the test set and learning curves on validation.

\subsection{Football -- \footballtask}
\label{sec:task-football}
\footballtask is a task introduced by Kurach~\etal~\cite{GoogleResearchFootball}. In this multi-agent task, three agents collaborate to score against rule-based defenders.
One agent starts behind the penalty arc and the other two agents start on the sides of the penalty area. There are two opponent rule-based defenders, including one goalkeeper. At the beginning of each episode, the center agent possesses the ball and faces the defender. The episode terminates when the controlled agents score or the maximum episode length is reached. 

\noindent\textbf{Shaped rewards.} We use the `checkpoint'~\cite{GoogleResearchFootball} reward as the goal-dependent reward $r^{\text{progress}}_t$ to encourage  controlled agents to move the ball towards the opponent's goal.
Specifically, the area between the initial location of the center agent and the opponent goal is divided into three checkpoint regions according to the distance to the opponent goal.
A progress reward of $+0.1$ is received the first time the controlled players posses the ball in a checkpoint region.   

\noindent\textbf{Terminal rewards.} Following  \cite{GoogleResearchFootball}, $r_{\text{success}}=1$ and $\alpha=0$.

\noindent\textbf{Model architecture.} For a fair comparison, we use the CNN architecture used by~\cite{GoogleResearchFootball, LiuNEURIPS2020}, and a parallel PPO algorithm~\cite{LiuNEURIPS2020} to train the agents.

\noindent\textbf{Evaluation.} Agents are evaluated using the average score obtained in test episodes. Consistent with Liu~\etal~\cite{LiuNEURIPS2020}, we set a training budget of $8$ million environment steps. Training curves and final metrics are reported on test episodes.

\begin{table*}[t]
\vspace{-0.2cm}
\centering
\resizebox{\textwidth}{!}{%
\begin{tabular}{l>{\columncolor{ColorPNav}}c>{\columncolor{ColorPNav}}c>{\columncolor{ColorPNav}}c>{\columncolor{ColorPNav}}c>{\columncolor{ColorFurnMove}}c>{\columncolor{ColorFurnMove}}c>{\columncolor{ColorFurnMove}}c>{\columncolor{ColorFurnMove}}c>{\columncolor{ColorFootball}}c>{\columncolor{ColorFootball}}c}

\toprule
\multicolumn{1}{c}{Tasks $\rightarrow$} & \multicolumn{4}{c}{\cellcolor{ColorPNav}\textbf{PointGoal Navigation}} & \multicolumn{4}{c}{\cellcolor{ColorFurnMove}\textbf{Furniture Moving}} & \multicolumn{2}{c}{\cellcolor{ColorFootball}\textbf{3 \vs 1 with Keeper}} \\
\multicolumn{1}{c}{} & \multicolumn{2}{c}{\cellcolor{ColorPNav}SPL} & \multicolumn{2}{c}{\cellcolor{ColorPNav}Success} & \multicolumn{2}{c}{\cellcolor{ColorFurnMove}MD-SPL} & \multicolumn{2}{c}{\cellcolor{ColorFurnMove}Success} & \multicolumn{2}{c}{\cellcolor{ColorFootball}Game Score} \\
Training routines $\downarrow$ & \emph{@10\%} & \emph{@100\%} & \emph{@10\%} & \emph{@100\%} & \emph{@10\%} & \emph{@100\%} & \emph{@10\%} & \emph{@100\%} & \emph{@10\%} & \emph{@100\%} \\
\midrule
\directrl & 35.9 & 54.7 & 60.5 & 79.0 & 2.8 & 11.2 & 22.5 & 58.4 & 0.0 & 0.6 \\
\gtop & \textbf{57.6} & 69.0 & \textbf{77.5} & \textbf{86.4} & \textbf{8.4} & 9.7 & \textbf{57.7} & 62.0 & 0.37 & \textbf{0.65} \\
\gtop $\rightarrow$ \directrl & 53.6 & \textbf{69.1} & 71.9 & 84.9 & 7.1 & \textbf{15.3} & 55.3 & \textbf{68.6} & \textbf{0.38} & 0.61 \\
\midrule
Gridworld expert (\textit{upper bound}) & \multicolumn{2}{c}{\cellcolor{ColorPNav}\textit{85.3}} & \multicolumn{2}{c}{\cellcolor{ColorPNav}\textit{97.5}} & \multicolumn{2}{c}{\cellcolor{ColorFurnMove}\textit{22.2}} & \multicolumn{2}{c}{\cellcolor{ColorFurnMove}\textit{76.3}} & \multicolumn{2}{c}{\cellcolor{ColorFootball}\textit{0.95}} \\
\bottomrule
\end{tabular}%
}
\vspace{0.03in}
\caption{\textbf{Quantitative results (shaped reward structure).} 
Identical to \tabref{tab:terminal-metrics} but all methods have been trained with shaped, rather than terminal, rewards. Even with this denser form of supervision, training routines based on \gtop can improve performance of all the metrics across the three tasks.
}
\label{tab:dense-metrics}
\end{table*}

\begin{figure*}[t]
\vspace{-0.3cm}
    \centering
    \begin{tabular}{ccc}
        \multicolumn{3}{c}{\includegraphics[width=0.7\textwidth]{figs/legend_no_variants.pdf}}\vspace{-2mm}\\
        \includegraphics[trim=5 5 10 10,clip,width=0.3\textwidth]{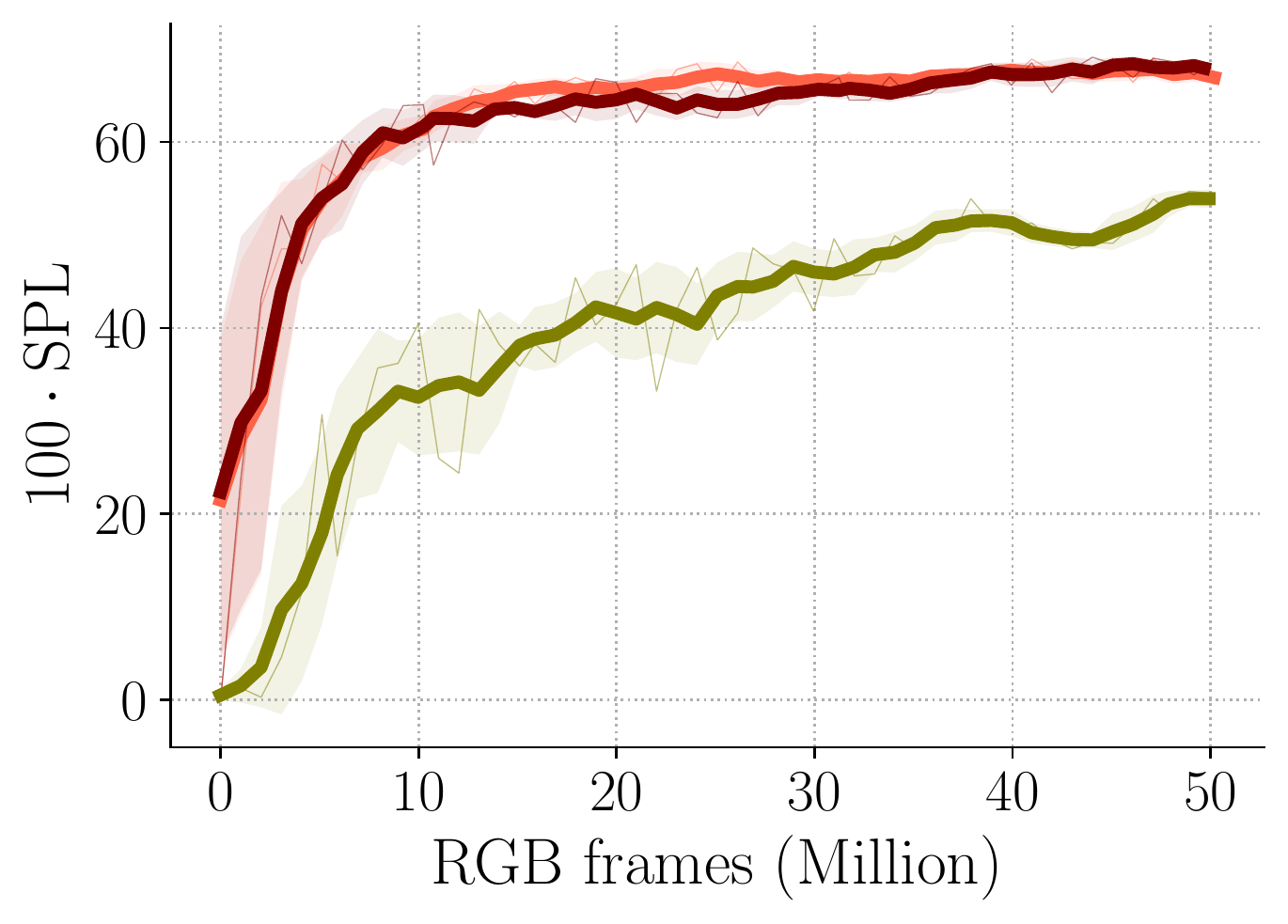} &
        \includegraphics[trim=5 5 10 10,clip,width=0.3\textwidth]{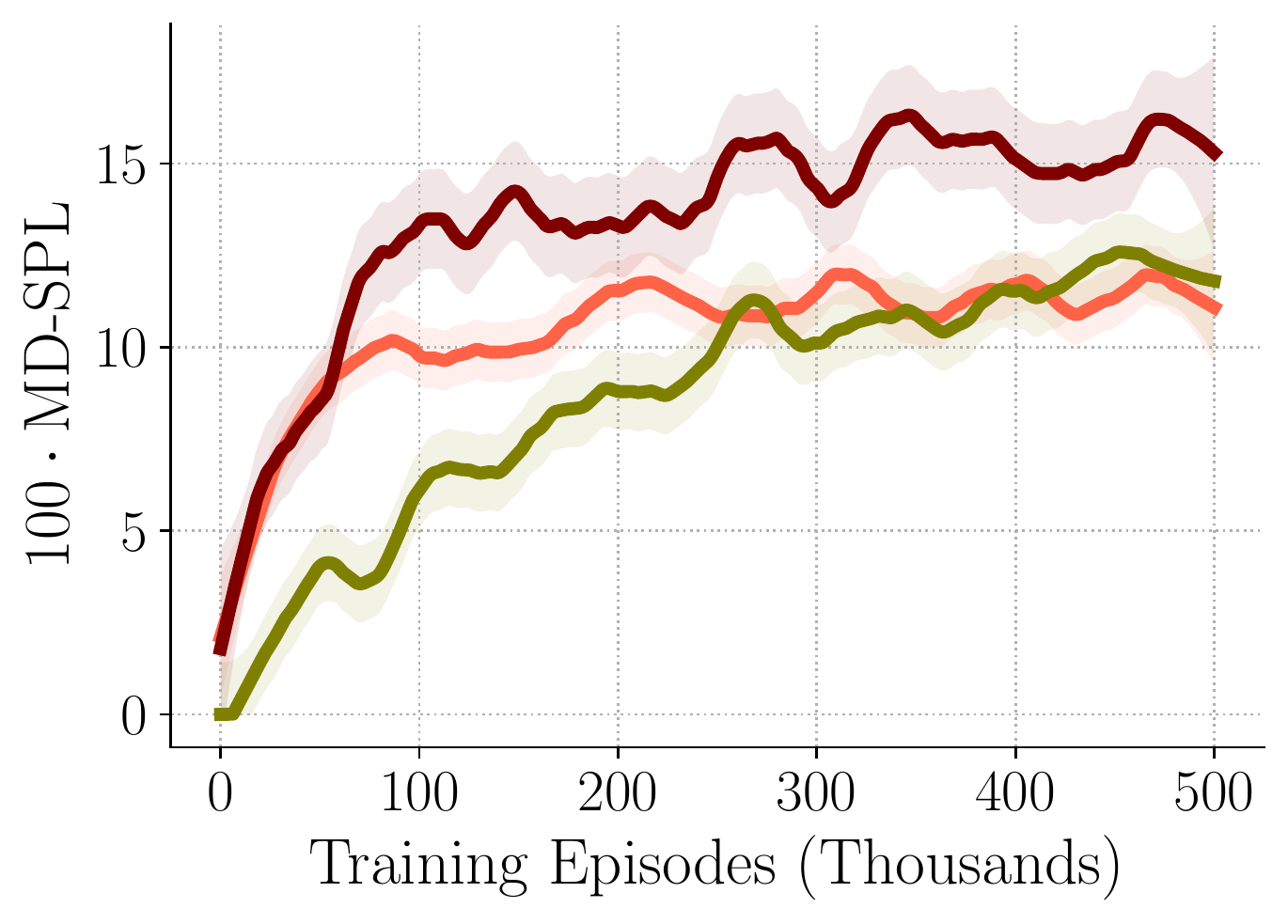} &
        \includegraphics[width=0.3\textwidth]{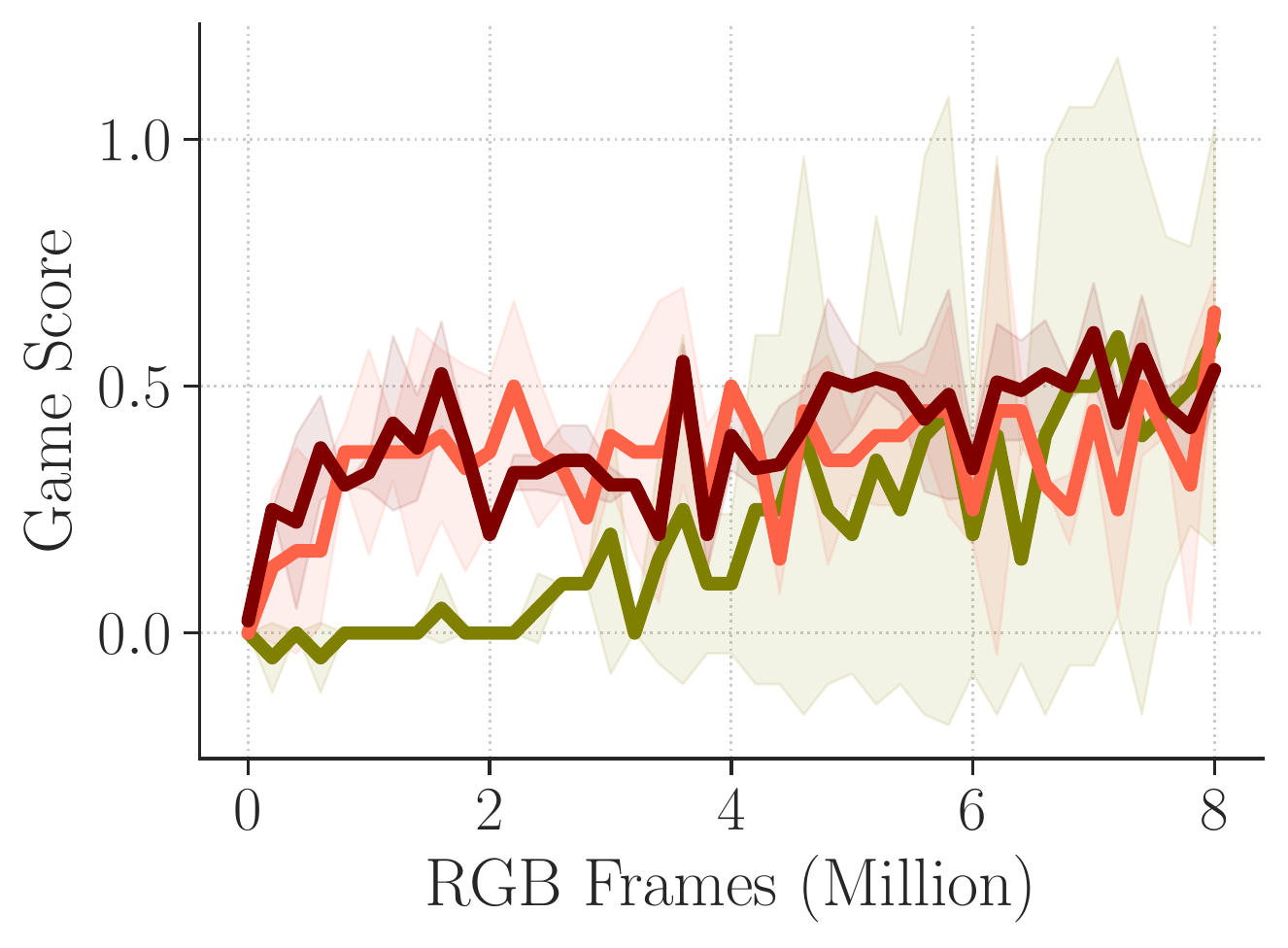}\\
        (a) PointGoal Navigation & (b) Furniture Moving & (c) Football -- 3 \vs 1 with Keeper
    \end{tabular}
    \vspace{0.03in}
    \caption{\textbf{Learning curves on validation set (shaped reward structure).} Follows~\figref{fig:terminal-plots}. As expected, with this additional supervision the performance gap between \gtop and \Directrl methods narrows but can still be substantial (\eg, in the \pnav task). See~\figref{fig:terminal-plots} for legend and plot details.
    \vspace{-0.5cm}
    }
    \label{fig:dense-plots}
\end{figure*}

\section{Experiments}
\label{sec:results}

We now provide an overview of the training routines employed to train our visual and gridworld agents, followed by results on the three tasks.

\subsection{Training Visual Agents}
\label{sec:training-routines-visual}
\noindent For each task (in both the terminal reward and shaped reward settings), we train visual agents in three ways:\\
\noindent\textbf{\Directrl.} We train visual agents (hence, `Pix') using only reward-based RL losses (\ie, PPO or A3C).\\
\noindent\textbf{GridToPix.} Our proposed routine trains visual agents by imitating a corresponding gridworld expert (that we train). As stated before, the expert is only available at training time and  weights are fixed. 
See  appendix for more details.\\
\noindent\textbf{GridToPix$\rightarrow$\Directrl.} This is a hybrid of the above two routines and follows the common practice of `warm-starting' agent policies with supervised/imitation learning and then fine-tuning with reinforcement learning.\footnote{For example, prior works use IL $\rightarrow$ RL to train agents for visual dialog~\cite{visdial_rl}, embodied question answering~\cite{DasECCV2018}, vision-and-language navigation~\cite{wang2018look,wang2019reinforced,jain2019stay}, and emergent communication~\cite{Lowe2020}.}
Additional details are deferred to the appendix.

\subsection{Training Gridworld Experts} 

For each task, the \textit{gridworld expert} utilizes a training routine identical to \Directrl. Gridworld experts observe semantic top-down tensors or 1D state information, unlike the architectures that learn from raw pixels. Hence, we make minimal edits to the CNN that encodes the observations (see appendix for details).

As described in~\secref{sec:app}, training in gridworlds is very fast. Moreover, gridworlds can, in principle, be optimized to accelerate the environment transitions. Hence, for gridworld variants of each of the three tasks, we train models to near saturation. Importantly, due to difference in state space, models, and training time, the \textit{gridworld expert} rows in~\tabref{tab:terminal-metrics} and \tabref{tab:dense-metrics} shouldn't be compared to visual analogues. They serve as a loose upper-bound for the performance of the visual agents trained with \gtop.

\subsection{Results}
We report  standard evaluation metrics on three tasks. To study sample efficiency, we also show   learning curves.

\noindent\textbf{Terminal rewards} (see \tabref{tab:terminal-metrics} and \figref{fig:terminal-plots}). With perfect perception, gridworld experts can train to a high performance (\eg, 94\% success in \pnav) and guide the learning of visual agents. In sharp contrast, \directrl doesn't learn a meaningful policy in any of the tasks\footnote{We investigated the learned policy qualitatively in the \pnav task and found that the \directrl agent learns the (locally optimal) policy of executing STOP as its first action. We report best number after training over 15 configurations (20M steps each) by varying the entropy coefficient, number of PPO trainers, and random seeds. For \footballtask, the DirectPix agents shoot straight at the goal and are intercepted by the goal-keeper and sometimes pass to another player who fails to receive.} -- demonstrating present day methods' inability to learn from terminal rewards in these visual worlds. Our \gtop variants perform significantly better. For instance, at \pnav, \gtop obtains a respectable SPL of 0.638, inching towards the 0.788 obtained by the gridworld expert. \gtoptod sometimes produces small gains beyond \gtop. As evidenced by learning curves, \gtop learning is also efficient -- most gains come in the early few million steps (0.45 SPL, \ie, 70\% of the final performance in just 5M steps for \pnav). A similar pattern emerges in other tasks -- \directrl cannot learn  meaningful behaviors from terminal rewards with a 0.8\% success and 0.07 game score in \fmove and \footballtask, respectively. Our training routines achieve the best performance. 

In addition to these improvements in the terminal reward structure, we also investigate the training routines in (the more traditional) setting with shaped rewards.

\noindent\textbf{Shaped rewards} (see \tabref{tab:dense-metrics} and \figref{fig:dense-plots}). As seen in past works, \directrl performs well when provided with shaped rewards. Interestingly, \gtoptod also outperforms \directrl in this reward setting in the first two tasks and performs roughly as well on the third task -- an over 25\% relative gain in SPL (\pnav), an over 35\% relative gain in MD-SPL (\task), and an over 8\% relative gain in game score (Football). The efficiency of learning is also high with \gtoptod reaching 83\% of its final SPL in just 5M steps for the \pnav task. Since \pnav is single-agent navigation, we trained an agent using the optimal shortest-path actions via IL.  \gtop   outperforms this baseline with (very) dense supervision.\footnote{By imitating optimal actions, the visual agent can reach a SPL at (10\%, 100\%) = (0.301, 0.687) and success at (10\%, 100\%) = (35.5, 76.7).} 

These results across three tasks running in three  simulators demonstrate the potential of \gtop  as we move towards training with terminal rewards. In addition \gtop also provides meaningful gains in our current Embodied AI training setups that employ shaped rewards.

%% file: sections/02_related.tex
\section{Related Work}
\label{sec:rel}
\noindent\textbf{Embodied task completion.} Development of visually realistic simulators~\cite{Chang3DV2017Matterport,ai2thor,stanford2d3d,igibson,habitat19iccv} has led to tremendous progress for embodied
agents. Agents are trained for a variety of tasks including multiple variants of indoor navigation~\cite{robothor,habitat19iccv,igibson,AllenAct,zeng2021pushing,weihs2021visual,szot2021habitat}, question answering~\cite{DasCVPR2018,GordonCVPR2018,Wijmans2019EQAPhoto}, instruction following~\cite{anderson2018vision,fried2018speaker,wang2019reinforced,krantz2020beyond,ku2020room,singh2020moca}, adversarial gameplay~\cite{weihs2021learning,chen2020visual}, and multi-agent collaboration~\cite{jain2019CVPRTBONE,JainWeihs2020CordialSync,GoogleResearchFootball,liu2021cooperative,patel2021comon}. 
In this work we have focused on three diverse  tasks, particularly, PointNav~\cite{habitat19iccv}, {\fmove}~\cite{JainWeihs2020CordialSync}, and \footballtask~\cite{GoogleResearchFootball}. In contrast to past works that generally aim to maximize success rates at these tasks using any and all possible supervision, we focus on achieving high performance while using only terminal rewards.

\noindent\textbf{Sparse rewards.} Learning from sparse rewards has long been of interest within the RL community where research has, in large part, focused on visually simple environments. A number of approaches have been proposed: curiosity as intrinsic motivation for exploring diverse states~\cite{pathakICMl17curiosity}, using hindsight to reinterpret ``failed'' trajectories~\cite{AndrychowiczEtAl2017HindsightHER}, curriculum learning~\cite{Bengio2009,FlorensaEtAl2017}, self-play~\cite{Silver2017MasteringTG,Baker2019EmergentTU}, and learning to shape rewards~\cite{Trott2019KeepingYD,Nair2018VisualRL,Ghosh2019LearningAR}. Unlike these approaches, we are interested in enabling models to learn from terminal rewards in visually complex environments. Our \gtop method can be further enhanced by incorporating any of the above ideas: rather than using standard RL to train our gridworld experts we could instead use one of the above methods and, potentially, obtain even better results. As our gridworld experts already learn high quality policies without these methods we leave this for future work.

\noindent\textbf{Accelerating visual reinforcement learning.} Several methods have been proposed to improve the sample and wall-clock efficiency of RL in embodied tasks. For sample efficiency, agents are often optimized with additional auxiliary self-supervised tasks~\cite{zhu2020vision,ye2020auxiliary}. 
For faster training, advances for distributed and decentralized scaling of PPO have been proposed~\cite{wijmans2019dd,LiuNEURIPS2020}. Pertinent to our work, Jain~\etal~\cite{JainWeihs2020CordialSync} developed an \thor-aligned gridworld ($16\times$ faster than \thor~\cite{ai2thor}) to prototype their experiments and scale their tasks to a larger number of agents.

\noindent\textbf{Imitation learning.}
An alternative to reward-based training of agents is imitation of a supervisory policy or demonstrations via behavior cloning~\cite{sammut1992learning,bain1995framework}. \Eg, Data Aggregation (DAgger)~\cite{ross2010efficient,RossAISTATS2011} mitigates the \textit{covariate-shift} that troubles classical behaviour cloning approaches. 
This has been applied to train visual agents via rule-based shortest path experts~\cite{GuptaCVPR2017,DasECCV2018,jain2019CVPRTBONE,WeihsJain2020Bridging}. 
\camera{
For the task of instruction following~\cite{shridhar2020alfred}, separate neural modules for perception and action have shown to improve policies learned by imitating human-annotated data~\cite{singh2020moca}.
Chen~\etal~\cite{chen2020learning} train gridworld agents to predict waypoints for AI driving.
Their privileged agents are trained by
offline behavior cloning of human-labeled trajectories (with
data augmentation). 
Obtaining human-labeled data for dense supervision for sequence prediction can quickly become costly and time-consuming. In contrast, we train visual agents using self-supervised labels from their gridworld counterparts. Since these experts learn by interacting in the gridworld via \textit{minimal supervision} of terminal rewards, no human-labeled data is needed.
}

%% file: sections/06_conclusion.tex
\section{Conclusion} \label{sec:conc}
Embodied AI progress is slowed by  labor-intensive work %
to manually tweak model architectures, collect human annotations, and/or shape reward functions. We study how  to reduce this effort significantly: for an embodied environment of interest, create a generic gridworld mirror which is visually parsimonious and fast to simulate. Use of this gridworld mirror for \gtop enables one to avoid many task-specific interventions and allows for learning from only terminal rewards: a critical step for complex tasks within physically realistic  environments where generating shaped rewards quickly becomes infeasible.

\noindent\textbf{Acknowledgements:} This work is supported in part by NSF under Grant \#1718221, 2008387, 2045586, MRI \#1725729, and NIFA award 2020-67021-32799. We thank Anand Bhattad, Angel X.\ Chang, Manolis Savva, Martin Lohmann, and Tanmay Gupta for thoughtful discussions and valuable input.

%% file: sections/07_supplementary.tex
\onecolumn
\section{Appendix of \textsc{GridToPix}: Training Embodied Agents with Minimal Supervision}\label{sec:supp}
\setcounter{table}{0}
\renewcommand{\thetable}{\Alph{section}.\arabic{table}}
\setcounter{figure}{0}
\renewcommand{\thefigure}{\Alph{section}.\arabic{figure}}

\noindent In this appendix we include:
\begin{enumerate}\compresslist
    \item[\ref{sec:supp-rewards}] Details of reward structure for \{\pnav, \task, \footballtask\}$\times$\{shaped, terminal\} (\tabref{tab:supp-reward}).
    \item[\ref{sec:supp-reward-2}] An alternate terminal reward structure for \task by simply dropping a component from $r^{\indep}_{t}$ (\tabref{tab:supp-furnmove-reward-struc-2}).
    \item[\ref{sec:supp-more-furnmove-metrics}] Additional quantitative metrics for \task introduced in~\cite{JainWeihs2020CordialSync} (terminal rewards -- \tabref{tab:supp-furnmove-terminal} and shaped rewards -- \tabref{tab:supp-furnmove-shaped}).
    \item[\ref{sec:supp-cnn-encoder}] Edits to allow deep net models to encode gridworld observation (\vs visual observations).
    \item[\ref{sec:supp-training-routines}] Teacher forcing probability and IL $\rightarrow$ RL transition for \gtop and \gtoptod routines (\tabref{tab:supp-training-routines}).
    \item[\ref{sec:supp-visual-mpe}] \camera{Experimental results for Visual Predator-Prey task in OpenAI multi-particle environment~\cite{LoweNIPS2017,MordatchAAAI2018}. This serves as a testbed wherein \textit{reward shaping is not enough} (\figref{fig:visual_mpe} and \tabref{tab:visual_pp_n_3}).}
    \item[\ref{sec:supp-qual-directrl}] Qualitative visualizations of policy learned by \Directrl agent in \footballtask (\figref{fig:supp-football-fail-passes} and \figref{fig:supp-football-fail-shoots}).
\end{enumerate}

\subsection{Reward Structures}
\label{sec:supp-rewards}
In~\tabref{tab:supp-reward}, we list all the reward components for shaped and terminal rewards for the three tasks considered in this work. For a fair comparison, the shaped rewards are kept identical to prior work in \pnav~\cite{habitat19iccv}, \task~\cite{JainWeihs2020CordialSync}, and \footballtask~\cite{GoogleResearchFootball}. For terminal rewards, as per the definitions in~\secref{sec:terminal-vs-shaped-rewards}, there exists no $r_t^{\text{progress}}$ component (hence, `\textcolor{red}{\ding{56}}'). The results for these reward structures have been reported in the main paper (\tabref{tab:terminal-metrics} and \tabref{tab:dense-metrics}).

Common across all tasks, $\mathcal{I}[\text{success}]$ denotes the indicator function conditioned on success of the episode. `Step penalty' is a small negative reward to encourages completion of the episode in fewer steps. In \pnav, the progress reward is based on the shortest-path geodesic distance to the goal from the current location of the agent ($d^{\text{geodesic}}_t$). In \task, particularly in~\cite{JainWeihs2020CordialSync}, the authors use a variant of the \pnav progress reward. Particularly, Manhattan distances are used ($d^{\text{Man.}}_t$) and a positive reward is received only if the agents get the furniture item  closer to the goal compared to the minimum distance in previous steps (note the use of `$\max \min$'). For \footballtask, the progress reward is called the `checkpoint reward'. It is received if the agents are able to move the ball to a zone closer to the goal.

\begin{table*}[h]
\centering
\resizebox{\textwidth}{!}{%
\begin{tabular}{lllll}
\toprule
\textbf{Task} & \textbf{Reward structure} & \textbf{$r^{\text{success}}_t$} & \textbf{$r^{\text{progress}}_t$} & \textbf{$r^{\indep}_t$} \\ \hline
\rowcolor{ColorPNav}
\pnav & Shaped & $10 \cdot \mathcal{I}[\text{success}]$ & $d^{\text{geodesic}}_{t-1} - d^{\text{geodesic}}_{t}$ & Step penalty $(-0.01)$ \\ \hline
\rowcolor{ColorPNav}
\pnav & Terminal & $10 \cdot ( 1 - 0.9 \cdot \frac{t}{T} ) \cdot \mathcal{I}[\text{success}]$ & \textcolor{red}{\ding{56}} & $0$ \\ \hline
\rowcolor{ColorFurnMove}
\task & Shaped & $1 \cdot \mathcal{I}[\text{success}]$ & \begin{tabular}[c]{@{}l@{}}$\max($\\$\min_{k=0, \ldots, t-1 }d^{\text{Man.}}_{k} - d^{\text{Man.}}_{t},$\\$0)$\end{tabular} & \begin{tabular}[c]{@{}l@{}}Failed action penalty $(-0.02)$\\ Failed coordination $(-0.1)$\\ Step penalty $(-0.01)$\end{tabular} \\ \hline
\rowcolor{ColorFurnMove}
\task & Terminal & $1 \cdot \mathcal{I}[\text{success}]$ & \textcolor{red}{\ding{56}} & \begin{tabular}[c]{@{}l@{}}Failed action penalty $(-0.02)$\\ Failed coordination $(-0.1)$\\ Step penalty $(-0.01)$\end{tabular} \\ \hline
\rowcolor{ColorFootball}
\footballtask & Shaped & $1 \cdot \mathcal{I}[\text{success}]$ & Checkpoint reward $(0.1)$& $0$ \\ \hline
\rowcolor{ColorFootball}
\footballtask & Terminal & $1 \cdot \mathcal{I}[\text{success}]$ & \textcolor{red}{\ding{56}} & $0$ \\\bottomrule
\end{tabular}%
}
\vspace{1mm}
\caption{\textbf{Reward structures.} For each task (\pnav, \task, and \footballtask), we list the three components of shaped and terminal reward structures. This includes a positive reward conditioned on success $r_t^{\text{success}}$, a goal-dependent progress reward $r_t^{\text{progress}}$, and a goal-independent reward $r^{\indep}_t$. Terminal rewards, that we focus on in this work, do not include progress reward (see~\secref{sec:terminal-vs-shaped-rewards} for definitions). Hence, progress reward is marked with a `\textcolor{red}{\ding{56}}' for terminal settings.}
\label{tab:supp-reward}
\end{table*}

\subsection{Alternate Terminal Reward Structures}
\label{sec:supp-reward-2}
\begin{wraptable}{r}{0.4\textwidth}%
\vspace{-10mm}
\centering
\resizebox{0.4\textwidth}{!}{%
\begin{tabular}{l>{\columncolor{ColorFurnMove}}c>{\columncolor{ColorFurnMove}}c}
\toprule
\textbf{Training Routine} & \textbf{MD-SPL} & \textbf{Success}\\\hline
\directrl & 0.1 & 2.5 \\
\gtop & 6.8 & 44.5  \\
\gtop $\rightarrow$ \directrl & 3.4 & 18.6 \\\hline
Gridworld expert (\textit{upper bound}) & \textit{17.0} & \textit{66.8} \\
\bottomrule
\end{tabular}%
}
\vspace{1mm}
\caption{Quantitative results for terminal rewards without failed action penalty (\task).}\vspace{-3mm}
\label{tab:supp-furnmove-reward-struc-2}
\end{wraptable}
As shown in~\tabref{tab:supp-reward}, the terminal reward structure for \task is equal to the shaped reward structure~\cite{JainWeihs2020CordialSync} except that it does not include the progress reward component $r^{\text{progress}}_t$. All results in the main paper (\task column of \tabref{tab:terminal-metrics} and \tabref{tab:dense-metrics}) and \tabref{tab:supp-furnmove-terminal} and \tabref{tab:supp-furnmove-shaped} correspond to this terminal reward setting. 

We, additionally, study an alternative terminal reward formulation where we drop the failed action penalty from the reward structure, \ie, only step penalty and failed coordination constitute $r^{\indep}_t$. We found the metrics to improve with this reward structure, with only 250k episode of training. These results are summarized in~\tabref{tab:supp-furnmove-reward-struc-2}. By further dropping the failed coordination penalty, the training became  sample-inefficient. This is coherent with the findings of Jain \etal~\cite{JainWeihs2020CordialSync}. Particularly, given the \textit{tightly-coupled} nature of the agents in collaboratively moving furniture, the failed coordination reward is critical for efficient learning.

\subsection{Additional \task Metrics}
\label{sec:supp-more-furnmove-metrics}
In the main paper, we report the primary metrics of \% successful episodes (Success) and a Manhattan distance based SPL (MD-SPL). For a fair comparison, we report three additional metrics that were included in~\cite{JainWeihs2020CordialSync}. This includes number of actions taken per agent (\textit{Ep Length}), probability of uncoordinated actions (\textit{Invalid Prob Mass}), and meters from goal at the episode's end (\textit{Final Distance}). \tabref{tab:supp-furnmove-terminal} and \tabref{tab:supp-furnmove-shaped} supplement the results reported in \tabref{tab:terminal-metrics} and \tabref{tab:dense-metrics}, respectively. 

\begin{table*}[h]
\centering
\resizebox{0.9\textwidth}{!}{%
\begin{tabular}{l>{\columncolor{ColorFurnMove}}c>{\columncolor{ColorFurnMove}}c>{\columncolor{ColorFurnMove}}c>{\columncolor{ColorFurnMove}}c>{\columncolor{ColorFurnMove}}c}
\toprule
\textbf{Training Routine} & \textbf{MD-SPL} $\uparrow$& \textbf{Success} $\uparrow$& \textbf{Ep Length} $\downarrow$& \textbf{Invalid Prob Mass} $\downarrow$& \textbf{Final Distance} $\downarrow$\\\hline
\directrl & 0.0 & 0.8 & 249.3 & 0.150 & 3.42 \\
\gtop & \textbf{4.0} & \textbf{24.6} & \textbf{210.7} & \textbf{0.110} & \textbf{2.848} \\
\gtop $\rightarrow$ \directrl & 3.1 & 14.5 & 224.2 & 0.116 & 3.215 \\\hline
Gridworld expert (\textit{upper bound}) & \textit{19.2} & \textit{56} & \textit{139.0} & \textit{0.077} & \textit{1.943}\\
\bottomrule
\end{tabular}%
}
\vspace{1mm}
\caption{\textbf{Additional quantitative rewards (terminal reward structure).} This table supplements results reported in~\tabref{tab:terminal-metrics}. In addition to metrics of MD-SPL and success rate, we include other relevant metrics. Vertical arrows, \ie, $\uparrow$ and $\downarrow$ denotes whether larger or smaller metric values are preferred, respectively.}
\label{tab:supp-furnmove-terminal}
\end{table*}

\begin{table*}[h]
\centering
\resizebox{0.9\textwidth}{!}{%
\begin{tabular}{l>{\columncolor{ColorFurnMove}}c>{\columncolor{ColorFurnMove}}c>{\columncolor{ColorFurnMove}}c>{\columncolor{ColorFurnMove}}c>{\columncolor{ColorFurnMove}}c}
\toprule
\textbf{Training Routine} & \textbf{MD-SPL} $\uparrow$& \textbf{Success} $\uparrow$& \textbf{Ep Length} $\downarrow$& \textbf{Invalid Prob Mass} $\downarrow$& \textbf{Final Distance} $\downarrow$\\\hline
\directrl & 11.2 & 58.4 & 155.7 & 0.311 & 1.154 \\
\gtop & 9.7 & 62.0 & 154.6 & 0.264 & 1.17 \\
\gtop $\rightarrow$ \directrl & \textbf{15.3} & \textbf{68.6} & \textbf{133.6} & \textbf{0.213} & \textbf{0.826} \\\hline
Gridworld expert (\textit{upper bound}) & \textit{22.2} & \textit{76.3} & \textit{109.7} & \textit{0.275} & \textit{0.722}\\
\bottomrule
\end{tabular}%
}
\vspace{1mm}
\caption{\textbf{Additional quantitative rewards (shaped reward structure).} This table supplements results reported in~\tabref{tab:dense-metrics}. }
\label{tab:supp-furnmove-shaped}
\end{table*}

\subsection{Gridworld Encoder and Implementation Details}
\label{sec:supp-cnn-encoder}
The models utilized for visual and gridworld agents are alike. Particularly, we make minimal edits to adapt the observation encoder to be able to process gridworld tensors instead of RGB tensors. This is briefly summarized below.

\noindent\textbf{\pnav.} For the \pnav, the visual encoder transforms a $(3, 256, 256)$ RGB tensor into a feature of length $512$ via three convolutional blocks and a linear layer. The grid encoder transforms a $(1, 100, 100)$ top-down tensor into a feature of the same length as the visual counterpart ($512$) via four convolutional blocks and a linear layer. For both visual and gridworld agents, the policy is represented via a GRU of hidden size $512$ followed by linear layers to serve as actor and critic heads. 

\noindent\textbf{\fmove.} Similarly as for \pnav, for \fmove the visual encoder transforms a $(3, 84, 84)$ RGB tensor into a feature of length $512$ via five convolutional blocks (no linear layers). The grid encoder transforms a $(9, 15, 15)$ top-down tensor into a feature of length $512$ via four convolutional blocks (for consistency, no linear layers). For both visual and gridworld agents in \task task, the policy is represented via a LSTM of hidden size $512$ followed by linear layers to serve as actor and critic heads. Note that we use the (best-performing) mixture-of-marginals actor head introduced in~\cite{JainWeihs2020CordialSync} for all \task experiments.

\noindent\textbf{\footballtask.} For the \footballtask task, the $(3, 1280, 960)$ RGB tensor is scaled to a $(1, 96, 72)$ gray-scale image. The visual encoder transforms this $(1, 96, 72)$ tensor into a feature of length $512$ via three convolution layers and a linear layer. The gridworld model observes the state as a vector of length $572$ and transforms it to a feature of length $64$ using a three linear layers. For both visual and gridworld agents, linear layers on top of the extracted feature serve as actor and critic heads.

\subsection{Training Routines}
\label{sec:supp-training-routines}
In the main paper we compare our \gtop and \gtoptod routines with \Directrl (\secref{sec:training-routines-visual}). Here, we include different exploration policies for imitation learning. We also include additional details of teacher forcing~\cite{supp-williams1989learning,supp-Bengio2015ScheduledSF} and IL $\rightarrow$ RL transition (\gtoptod).

\camera{\noindent\textbf{Exploratory policies $\mu$.}} Recall, from \secref{sec:super-vis-agents-via-grid}, the \gtop loss is
\begin{equation}
    \mathcal{L}_{\text{\gtop}} = \mathbb{E}[\mathbb{E}_{a\sim\mg(\cdot \mid O_\mu, H_\mu)}[-\log\mv(a\mid O_\mu, H_\mu)]]. 
\end{equation}
The choice of exploratory policy $\mu$ leads to three variants, each widely adopted in IL %
tasks: %
\\
\noindent$\bullet$ Student forcing (SF): This is an \textit{on-policy} method, \ie, the target policy $\mv$ to be learnt  is the exploration policy $\mu$. \\
\noindent$\bullet$ Teacher forcing (TF): In this case we use $\mu=\mg$, \ie, the visual agent takes the expert's actions. This helps the visual agent frequently observe states closer to the goal, which it would only see late  in training if following SF. The downside of TF: the visual agent $\av$ observes only states which meaningfully lead to the goal. Hence, TF is susceptible to \textit{covariate shift}, \ie, the visual agent  $\av$ exhibits low resilience to recover  
if it ventures `off-track.'\\
\noindent$\bullet$ Annealed teacher forcing or DAgger (DA): Agents take actions by combining SF and TF via $a_{t}^{\av}=(1-\beta) a_{t}^{\text{SF}} + \beta a_{t}^{\text{TF}}$ where $\beta\sim\text{Bernoulli}(p)$. A decay of $p$ from $1$ to $0$ is adopted during training to transition smoothly from TF to SF. After such annealing, the visual agent's policy $\mv$ is generally trained with SF until convergence.

\camera{\noindent\textbf{Teacher forcing probability}}. Details of teacher forcing for our methods are summarized in \tabref{tab:supp-training-routines}. We employ annealed teacher forcing (see~\secref{sec:super-vis-agents-via-grid}) for a part of the training budget. We anneal (decay) the teacher forcing probability linearly for \pnav, \task, and exponentially for the \footballtask. 

The \gtop routine is purely IL and is denoted in~\tabref{tab:supp-training-routines} with an IL arrow ({\color{orange}$\longleftrightarrow$}) spanning from 0\% to 100\%. In contrast, \gtoptod %
includes a warm-start with IL followed by reward-based learning. Hence, in~\tabref{tab:supp-training-routines}, we have a shorter IL arrow followed by an RL arrow ({\color{yellow!80!black}$\longleftrightarrow$}) along with the algorithm used to maximize rewards.

\begin{table*}[h]
    \centering
    \resizebox{0.9\textwidth}{!}{%
    \begin{tabular}{ccc}
        \toprule
        Task & \gtop & \gtoptod\\\hline
        \rowcolor{ColorPNav}
        \pnav
        &
        \includegraphics[valign=m,width=0.4\textwidth]{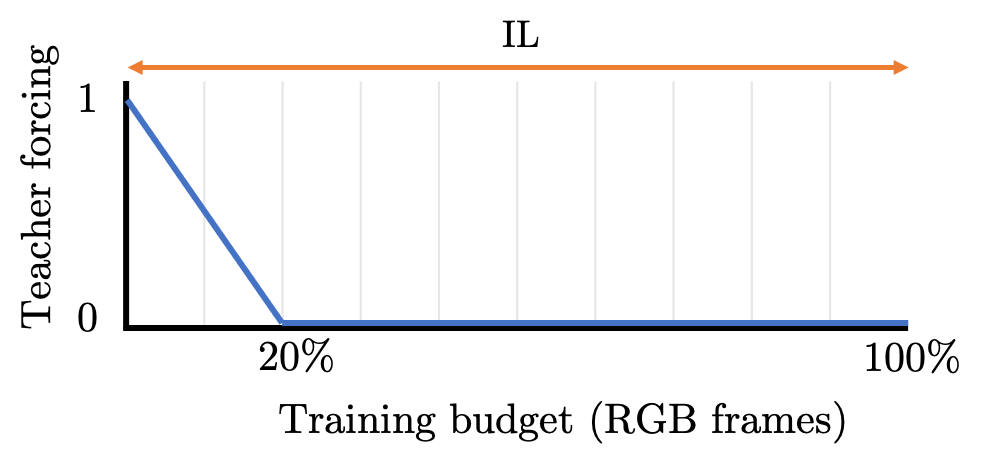} 
        &
        \includegraphics[valign=m,width=0.4\textwidth]{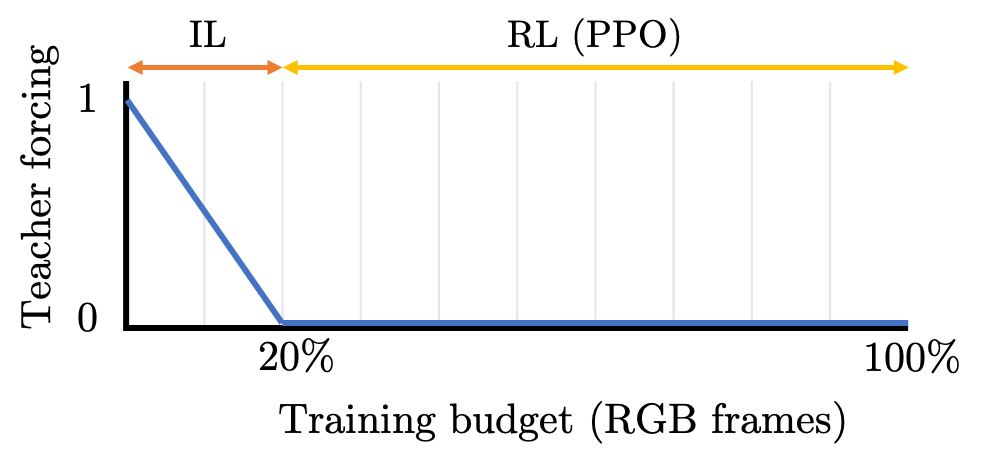}
        \\\hline
        \rowcolor{ColorFurnMove}
        \task
        &
        \includegraphics[valign=m,width=0.4\textwidth]{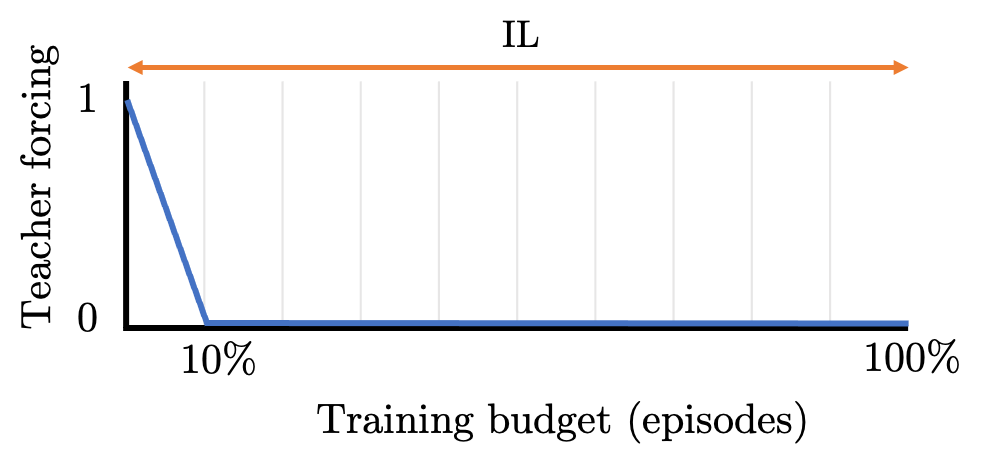} 
        &
        \includegraphics[valign=m,width=0.4\textwidth]{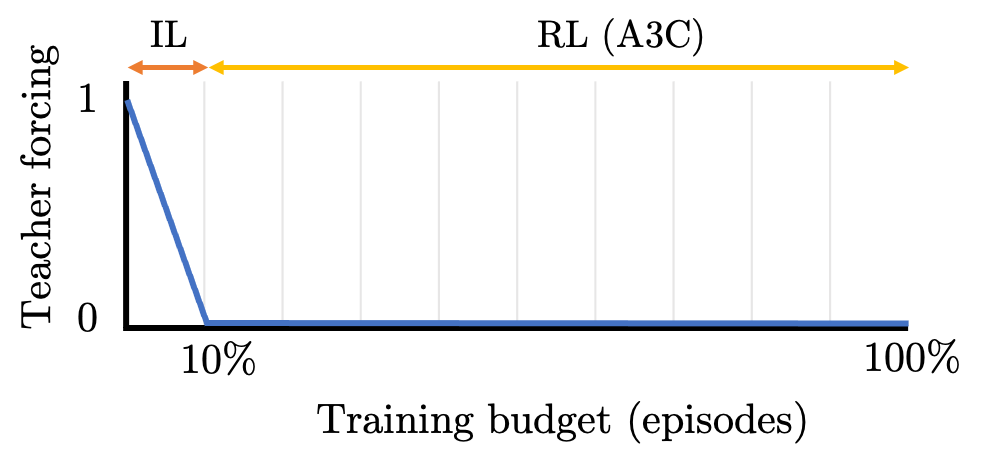}
        \\\hline
        \rowcolor{ColorFootball}
        \footballtask
        &
        \includegraphics[valign=m,width=0.4\textwidth]{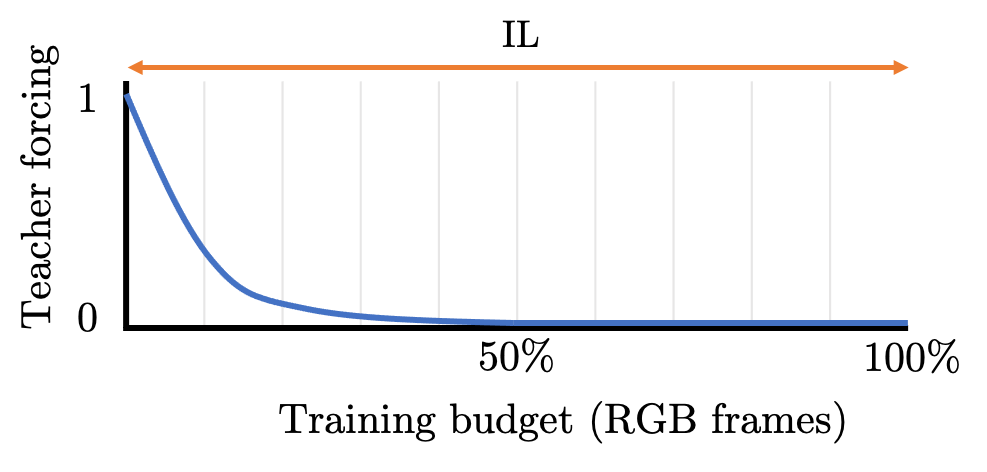} 
        &
        \includegraphics[valign=m,width=0.4\textwidth]{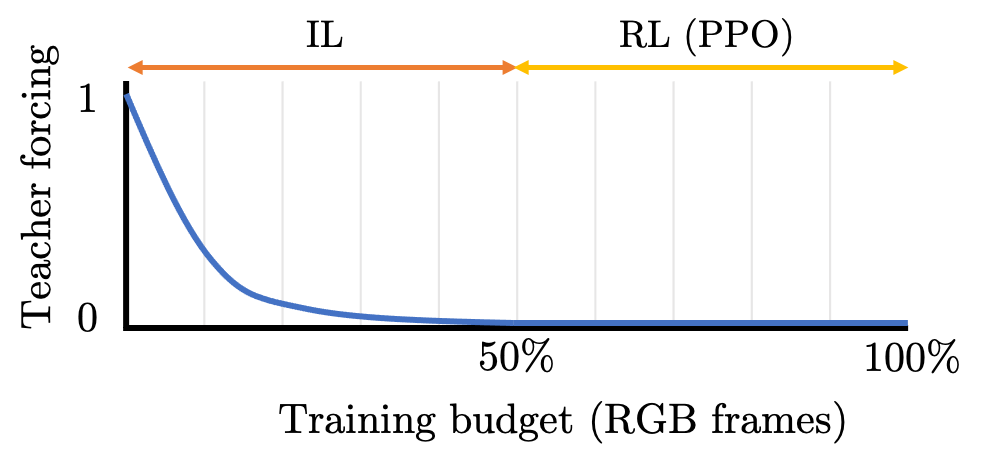}
        \\\bottomrule
    \end{tabular}
    }
    \vspace{1mm}
    \caption{\textbf{Training routines.} The figures show how teacher forcing probability varies over training and the transition from IL to RL.}
    \label{tab:supp-training-routines}
\end{table*}

\camera{\subsection{\mpe \ -- Task Where Reward Shaping Is Not Enough}
\label{sec:supp-visual-mpe}
Predator-prey is a multi-agent task defined within the OpenAI multiple-particle environments (MPE)~\cite{LoweNIPS2017}. It entails controlling a team of predators while a competing team of prey is controlled by the game engine. Specifically, $n$ predators work together to chase a team of $n/3$ prey that move faster than the predators. The objective is to optimize rewards by controlling the policies of the predators. 
Prior work~\cite{LoweNIPS2017, LiuCORL2019, JiangNIPS2018} assumes an agent observes a 1D vector summarizing the positions and velocities of all agents in a neighborhood. For consistency, we consider this 1D vector to be our \textit{gridworld} observation. In addition, akin to~\cite{liu2021iros}, we define an analogous \textit{visual} setting where agents only process a top-down map in pixel space. 

Note, for visual tasks like \pnav, \fmove, and \footballtask, we created (or leveraged existing) gridworlds. As standard visual tasks are mostly navigational, reward shaping is
typically tractable. Hence, to build a testbed where reward shaping isn't tractable, we are effectively creating the visual world for the complex, multi-agent predator-prey task.

Below, we include experimental details and corresponding results. Note, despite basic reward shaping, \directrl agents cannot learn whereas \gtop comes close to the upper bound of gridworld experts.

\noindent\textbf{Gridworld observation.}
The predator observes its location and velocity, the relative location of the neighboring landmarks and fellow predators, and the relative location and velocity of the three preys.

\noindent\textbf{Shaped rewards.}
Due to the complexity of the task,
it's intractable to \textit{perfectly} shape the reward. However, a positive reward for bumping into a prey and a negative reward based on the distance to the prey is provided. 

\noindent\textbf{Model architecture.}
We use a standard CNN model~\cite{MnihNature2015}. Our model has four hidden layers. The first three layers are convolutional layers with $32, 64,$ and $64$ filters. All convolutional layers have filters of size $4\times4$ and stride $2$. 
The fourth layer is a fully connected layer
with $256$ hidden units. Following the last hidden layer are linear layers that predict the value and actor policy. 

\noindent\textbf{Evaluation.}
Agents are evaluated using the average rewards obtained in test episodes. We train the agents for $3$M environment steps. Learning curves are reported on test episodes. 

\noindent\textbf{Results.} We experiment with $\text{number of predators} = n = \{3, 6\}$ visual predator-prey tasks. The learning curves for $n=3$ and setup for $n=6$ are illustrated in \figref{fig:visual_mpe}. The average rewards for \directrl, \gtop, \gtoptod, and gridworld experts are included in \tabref{tab:visual_pp_n_3}. Despite basic reward shaping, joint optimization of perception and planning leads to a \directrl policy demonstrating no meaningful behaviour. With the help of self-supervision via the gridworld expert, \gtop and \gtoptod perform significantly better with $48.7$ and $37.9$ average rewards over the training budget of $3$M steps. Results on the $n=6$ setting show a similar trend: \directrl and \gtop obtain average rewards of $-70.2$ and $237.1$ (the gridworld expert earns $281.1$).
\begin{figure}[t]
    \centering
    \includegraphics[width=.9\linewidth]{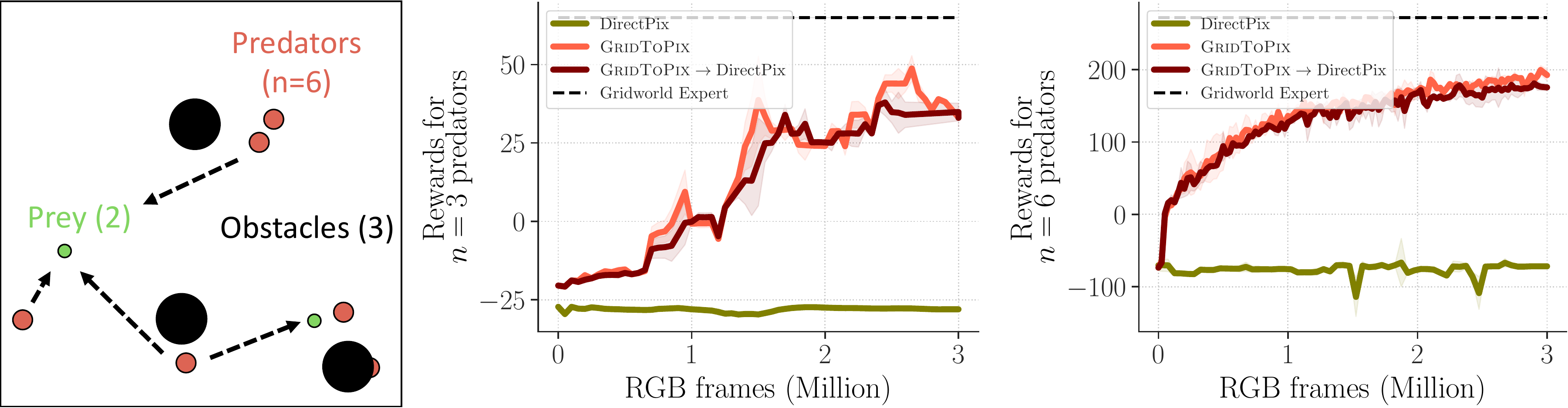}
    \caption{Visual Prey-Predator: task setup and learning curve.}
    \label{fig:visual_mpe}
    \vspace{-2mm}
\end{figure}

\begin{table}[t]
    \centering
    \resizebox{0.5\linewidth}{!}{%
    \begin{tabular}{ccccc}
        \toprule
        \multirow{2}{*}{\textbf{Method}} & \multicolumn{2}{c}{\textbf{Reward} ($n=3$)} & \multicolumn{2}{c}{\textbf{Reward} ($n=6$)}\\
         & \textit{@10\%} & \textit{@100\%} & \textit{@10\%} & \textit{@100\%} \\
         \midrule
        DirectPix & -27.6 & -28.0 & -71.2 & -70.2\\
        \gtop & -16.9 & 48.7 & 52.3  & 192.5\\
        \gtoptod & -17.1 & 37.9 & 47.3 & 175.5 \\
        \midrule
        Grid expert \textit{(upper bound)} & \multicolumn{2}{c}{65.0} & \multicolumn{2}{c}{281.1}\\
        \bottomrule
    \end{tabular}
    }
    \vspace{1mm}
    \caption{Visual \textit{predator-prey} for $\text{num. predators} = n = 3$.
    }
    \label{tab:visual_pp_n_3}
    \vspace{-2mm}
\end{table}

}

\subsection{Qualitative Results of \Directrl Trained with Terminal Rewards}
\label{sec:supp-qual-directrl}
\begin{figure}[h]
    \centering
    \includegraphics[width=0.8\textwidth]{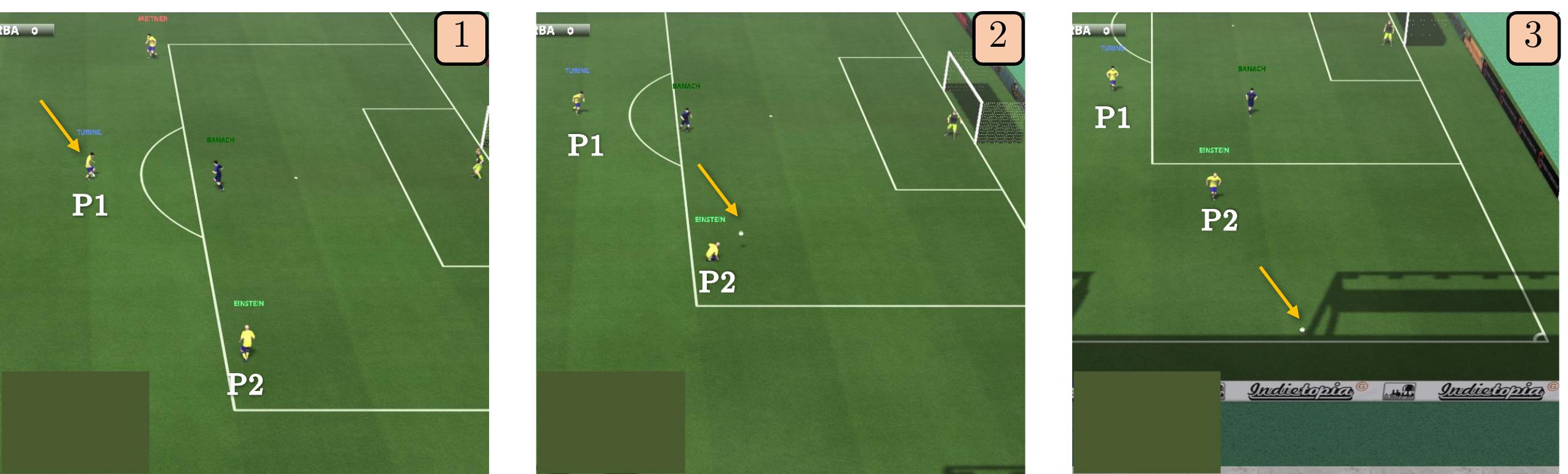}\\
    \includegraphics[width=0.8\textwidth]{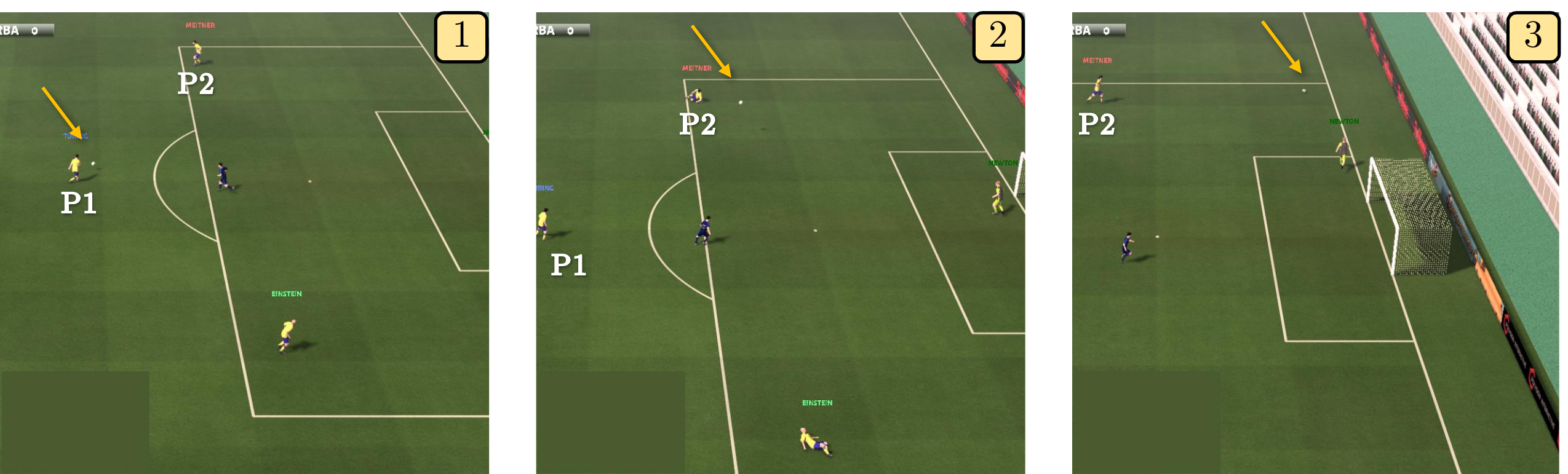}
    \caption{\textbf{Failure modes -- Misdirected passes.} Two episodes (top strip and bottom strip) that highlight a common failure mode of \Directrl training with terminal rewards. Particularly, player 1 attempts to pass the ball to player 2 but misdirects it to hit outside the field lines. For readability, the relevant players are marked as P1 and P2. Also, ball is highlighted using a yellow arrow.}
    \label{fig:supp-football-fail-passes}
\end{figure}
\begin{figure}[h]
    \centering
    \includegraphics[width=1\textwidth]{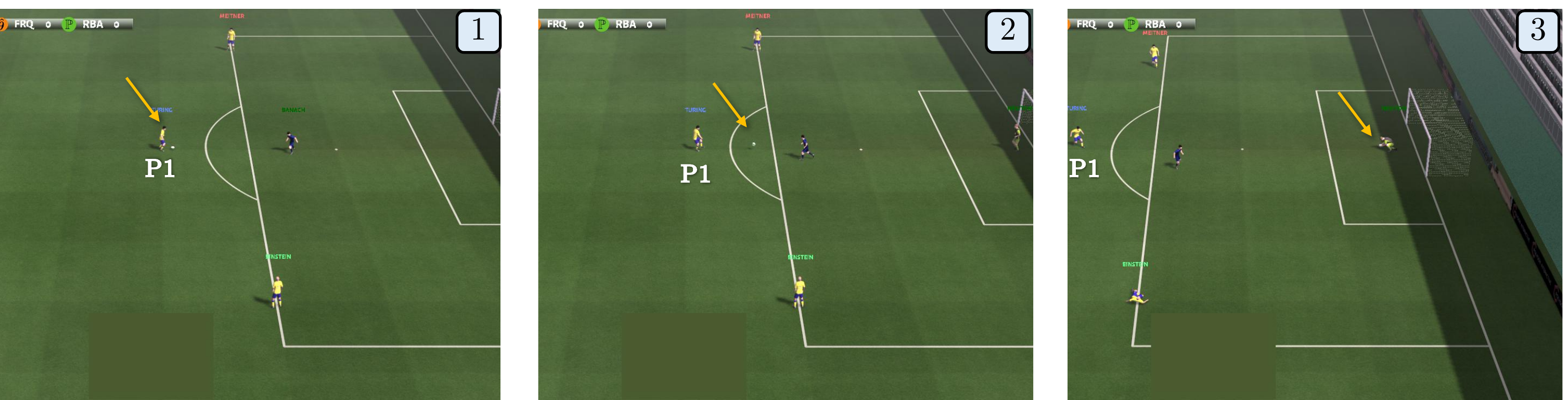}\\
    \includegraphics[width=1\textwidth]{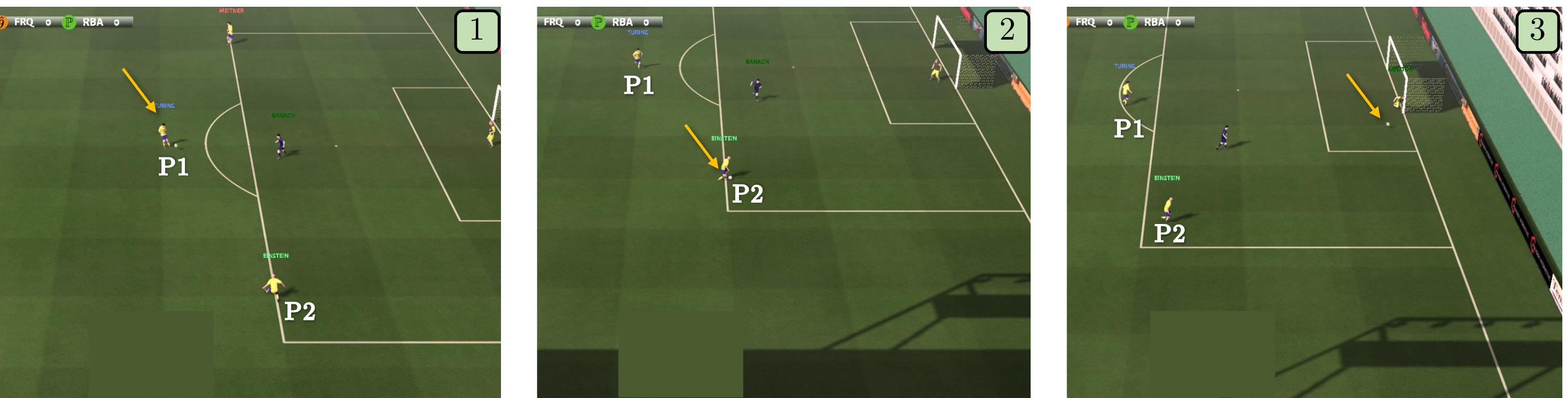}
    \caption{\textbf{Failure modes -- Long distance shooting.} Two episodes (top strip and bottom strip) that highlight a common failure mode of \Directrl training with terminal rewards. Particularly, in the top strip, player 1 attempts to score by shooting too ambitiously from the starting point. In the bottom strip, player 2 does the same, after receiving the ball from player 1. In both episodes the goal keeper can easily intercept the ball.}
    \label{fig:supp-football-fail-shoots}
\end{figure}

\label{sec:supp-directrl-visualization}
As we report in \secref{sec:results}, \Directrl doesn't learn a meaningful policy in any of the tasks when given only terminal rewards. Closer  inspection reveals that the \directrl agent for the \pnav task learns  a degenerate probability distribution with almost all probability mass allocated to the `Stop' action. Similarly, many of the strategies learned by \Directrl agents for \footballtask are also myopic. Particularly, the agents cannot effectively pass to each other, pushing the ball outside the field lines (see \figref{fig:supp-football-fail-passes}). Another common failure mode is shooting at the goal from too far off (see \figref{fig:supp-football-fail-shoots}).

%% file: main.bbl
\begin{thebibliography}{10}\itemsep=-1pt

\bibitem{anderson2018evaluation}
Peter Anderson, Angel Chang, Devendra~Singh Chaplot, Alexey Dosovitskiy,
  Saurabh Gupta, Vladlen Koltun, Jana Kosecka, Jitendra Malik, Roozbeh
  Mottaghi, Manolis Savva, et~al.
\newblock On evaluation of embodied navigation agents.
\newblock {\em arXiv preprint arXiv:1807.06757}, 2018.

\bibitem{anderson2018vision}
Peter Anderson, Qi Wu, Damien Teney, Jake Bruce, Mark Johnson, Niko
  S{\"u}nderhauf, Ian Reid, Stephen Gould, and Anton van~den Hengel.
\newblock Vision-and-language navigation: Interpreting visually-grounded
  navigation instructions in real environments.
\newblock In {\em CVPR}, 2018.

\bibitem{AndrychowiczEtAl2017HindsightHER}
Marcin Andrychowicz, Filip Wolski, Alex Ray, Jonas Schneider, Rachel Fong,
  Peter Welinder, Bob McGrew, Josh Tobin, OpenAI Pieter~Abbeel, and Wojciech
  Zaremba.
\newblock Hindsight experience replay.
\newblock In I. Guyon, U.~V. LuxbuAndrychowiczrg, S. Bengio, H. Wallach, R.
  Fergus, S. Vishwanathan, and R. Garnett, editors, {\em NeurIPS}, 2017.

\bibitem{stanford2d3d}
Iro Armeni, Sasha Sax, Amir~R Zamir, and Silvio Savarese.
\newblock Joint 2d-3d-semantic data for indoor scene understanding.
\newblock {\em arXiv preprint arXiv:1702.01105}, 2017.

\bibitem{bain1995framework}
Michael Bain and Claude Sammut.
\newblock A framework for behavioural cloning.
\newblock In {\em Machine Intelligence}, 1995.

\bibitem{Baker2019EmergentTU}
Bowen Baker, Ingmar Kanitscheider, Todor Markov, Yi Wu, Glenn Powell, Bob
  McGrew, and Igor Mordatch.
\newblock Emergent tool use from multi-agent autocurricula.
\newblock {\em arXiv preprint arXiv:1909.07528}, 2019.

\bibitem{supp-Bengio2015ScheduledSF}
S. Bengio, Oriol Vinyals, Navdeep Jaitly, and Noam Shazeer.
\newblock Scheduled sampling for sequence prediction with recurrent neural
  networks.
\newblock In {\em NeurIPS}, 2015.

\bibitem{Bengio2009}
Y. Bengio.
\newblock {Learning Deep Architectures for AI}.
\newblock {\em Foundations and Trends in ML}, 2009.

\bibitem{cartillier2020semantic}
Vincent Cartillier, Zhile Ren, Neha Jain, Stefan Lee, Irfan Essa, and Dhruv
  Batra.
\newblock Semantic mapnet: Building allocentric semanticmaps and
  representations from egocentric views.
\newblock {\em arXiv preprint arXiv:2010.01191}, 2020.

\bibitem{Chang3DV2017Matterport}
Angel Chang, Angela Dai, Thomas Funkhouser, Maciej Halber, Matthias Niessner,
  Manolis Savva, Shuran Song, Andy Zeng, and Yinda Zhang.
\newblock {Matterport3D}: Learning from {RGB-D} data in indoor environments.
\newblock In {\em 3DV}, 2017.

\bibitem{chang2020semantic}
Matthew Chang, Arjun Gupta, and Saurabh Gupta.
\newblock Semantic visual navigation by watching youtube videos.
\newblock In {\em NeurIPS}, 2020.

\bibitem{Chaplot2020Explore}
Devendra~Singh Chaplot, Saurabh Gupta, Abhinav Gupta, and Ruslan Salakhutdinov.
\newblock Learning to explore using active neural mapping.
\newblock In {\em ICLR}, 2020.

\bibitem{chen2020visual}
Boyuan Chen, Shuran Song, Hod Lipson, and Carl Vondrick.
\newblock Visual hide and seek.
\newblock In {\em Artificial Life Conference Proceedings}. MIT Press, 2020.

\bibitem{chen2019audio}
Changan Chen, Unnat Jain, Carl Schissler, Sebastia Vicenc~Amengual Gari, Ziad
  Al-Halah, Vamsi~Krishna Ithapu, Philip Robinson, and Kristen Grauman.
\newblock Soundspaces: Audio-visual navigation in 3d environments.
\newblock In {\em ECCV}, 2020.

\bibitem{chen2020learning}
Dian Chen, Brady Zhou, Vladlen Koltun, and Philipp Kr{\"a}henb{\"u}hl.
\newblock Learning by cheating.
\newblock In {\em CoRL}, 2020.

\bibitem{babyai_iclr19}
Maxime Chevalier-Boisvert, Dzmitry Bahdanau, Salem Lahlou, Lucas Willems,
  Chitwan Saharia, Thien~Huu Nguyen, and Yoshua Bengio.
\newblock Baby{AI}: First steps towards grounded language learning with a human
  in the loop.
\newblock In {\em ICLR}, 2019.

\bibitem{hog}
N. Dalal and B. Triggs.
\newblock Histograms of oriented gradients for human detection.
\newblock {\em CVPR}, 2005.

\bibitem{DasCVPR2018}
A. Das, S. Datta, G. Gkioxari, S. Lee, D. Parikh, and D. Batra.
\newblock {Embodied Question Answering}.
\newblock In {\em CVPR}, 2018.

\bibitem{DasECCV2018}
A. Das, G. Gkioxari, S. Lee, D. Parikh, and D. Batra.
\newblock {Neural Modular Control for Embodied Question Answering}.
\newblock In {\em ECCV}, 2018.

\bibitem{visdial_rl}
Abhishek Das, Satwik Kottur, Jos\'e~M.F. Moura, Stefan Lee, and Dhruv Batra.
\newblock Learning cooperative visual dialog agents with deep reinforcement
  learning.
\newblock In {\em ICCV}, 2017.

\bibitem{robothor}
Matt Deitke, Winson Han, Alvaro Herrasti, Aniruddha Kembhavi, Eric Kolve,
  Roozbeh Mottaghi, Jordi Salvador, Dustin Schwenk, Eli VanderBilt, Matthew
  Wallingford, Luca Weihs, Mark Yatskar, and Ali Farhadi.
\newblock {RoboTHOR: An Open Simulation-to-Real Embodied AI Platform}.
\newblock In {\em CVPR}, 2020.

\bibitem{FlorensaEtAl2017}
Carlos Florensa, David Held, Markus Wulfmeier, Michael Zhang, and Pieter
  Abbeel.
\newblock Reverse curriculum generation for reinforcement learning.
\newblock In {\em CoRL}, 2017.

\bibitem{fried2018speaker}
Daniel Fried, Ronghang Hu, Volkan Cirik, Anna Rohrbach, Jacob Andreas,
  Louis-Philippe Morency, Taylor Berg-Kirkpatrick, Kate Saenko, Dan Klein, and
  Trevor Darrell.
\newblock Speaker-follower models for vision-and-language navigation.
\newblock In {\em NeurIPS}, 2018.

\bibitem{Gan2020ThreeDWorldAP}
Chuang Gan, Jeremy Schwartz, Seth Alter, Martin Schrimpf, James Traer, Julian
  De~Freitas, Jonas Kubilius, Abhishek Bhandwaldar, Nick Haber, Megumi Sano,
  et~al.
\newblock Threedworld: A platform for interactive multi-modal physical
  simulation.
\newblock {\em arXiv preprint arXiv:2007.04954}, 2020.

\bibitem{Ghosh2019LearningAR}
Dibya Ghosh, Abhishek Gupta, and Sergey Levine.
\newblock Learning actionable representations with goal-conditioned policies.
\newblock {\em arXiv preprint arXiv:1811.07819}, 2018.

\bibitem{GordonCVPR2018}
D. Gordon, A. Kembhavi, M. Rastegari, J. Redmon, D. Fox, and A. Farhadi.
\newblock {IQA: Visual Question Answering in Interactive Environments}.
\newblock In {\em CVPR}, 2018.

\bibitem{GuptaCVPR2017}
S. Gupta, J. Davidson, S. Levine, R. Sukthankar, and J. Malik.
\newblock {Cognitive Mapping and Planning for Visual Navigation}.
\newblock In {\em CVPR}, 2017.

\bibitem{hahn2020you}
Meera Hahn, Jacob Krantz, Dhruv Batra, Devi Parikh, James~M. Rehg, Stefan Lee,
  and Peter Anderson.
\newblock Where are you? localization from embodied dialog.
\newblock In {\em EMNLP}, 2020.

\bibitem{JainWeihs2020CordialSync}
Unnat Jain, Luca Weihs, Eric Kolve, Ali Farhadi, Svetlana Lazebnik, Aniruddha
  Kembhavi, and Alexander~G. Schwing.
\newblock A cordial sync: Going beyond marginal policies for multi-agent
  embodied tasks.
\newblock In {\em ECCV}, 2020.

\bibitem{jain2019CVPRTBONE}
Unnat Jain, Luca Weihs, Eric Kolve, Mohammad Rastegari, Svetlana Lazebnik, Ali
  Farhadi, Alexander~G. Schwing, and Aniruddha Kembhavi.
\newblock Two body problem: Collaborative visual task completion.
\newblock In {\em CVPR}, 2019.

\bibitem{jain2019stay}
Vihan Jain, Gabriel Magalhaes, Alexander Ku, Ashish Vaswani, Eugene Ie, and
  Jason Baldridge.
\newblock Stay on the path: Instruction fidelity in vision-and-language
  navigation.
\newblock In {\em ACL}, 2019.

\bibitem{RLBench}
S. James, Z. Ma, David~Rovick Arrojo, and A. Davison.
\newblock Rlbench: The robot learning benchmark and learning environment.
\newblock In {\em IEEE Robotics and Automation Letters}, 2020.

\bibitem{JiangNIPS2018}
Jiechuan Jiang and Zongqing Lu.
\newblock Learning attentional communication for multi-agent cooperation.
\newblock In {\em NeurIPS}, 2018.

\bibitem{Sim2RealHabitat}
Abhishek Kadian, Joanne Truong, Aaron Gokaslan, Alexander Clegg, Erik Wijmans,
  Stefan Lee, Manolis Savva, Sonia Chernova, and Dhruv Batra.
\newblock Sim2real predictivity: Does evaluation in simulation predict
  real-world performance?
\newblock {\em {IEEE} Robotics Automation Letters}, 2020.

\bibitem{ai2thor}
Eric Kolve, Roozbeh Mottaghi, Winson Han, Eli VanderBilt, Luca Weihs, Alvaro
  Herrasti, Daniel Gordon, Yuke Zhu, Abhinav Gupta, and Ali Farhadi.
\newblock {AI2-THOR:} an interactive 3d environment for visual {AI}.
\newblock {\em arXiv preprint arXiv:1712.05474}, 2019.

\bibitem{krantz2020beyond}
Jacob Krantz, Erik Wijmans, Arjun Majumdar, Dhruv Batra, and Stefan Lee.
\newblock Beyond the nav-graph: Vision-and-language navigation in continuous
  environments.
\newblock In {\em ECCV}, 2020.

\bibitem{ku2020room}
Alexander Ku, Peter Anderson, Roma Patel, Eugene Ie, and Jason Baldridge.
\newblock Room-across-room: Multilingual vision-and-language navigation with
  dense spatiotemporal grounding.
\newblock In {\em EMNLP}, 2020.

\bibitem{GoogleResearchFootball}
Karol Kurach, Anton Raichuk, Piotr Sta{\'n}czyk, Micha{\l} Zaj{\k{a}}c, Olivier
  Bachem, Lasse Espeholt, Carlos Riquelme, Damien Vincent, Marcin Michalski,
  Olivier Bousquet, et~al.
\newblock Google research football: A novel reinforcement learning environment.
\newblock In {\em AAAI}, 2020.

\bibitem{Lazebnik2006BeyondBO}
S. Lazebnik, C. Schmid, and J. Ponce.
\newblock Beyond bags of features: Spatial pyramid matching for recognizing
  natural scene categories.
\newblock {\em CVPR}, 2006.

\bibitem{liu2021cooperative}
Iou-Jen Liu, Unnat Jain, Raymond~A Yeh, and Alexander Schwing.
\newblock Cooperative exploration for multi-agent deep reinforcement learning.
\newblock In {\em ICML}, 2021.

\bibitem{liu2021iros}
Iou-Jen Liu, Zhongzheng Ren, Raymond~A. Yeh, and Alexander~G. Schwing.
\newblock Semantic tracklets: An object-centric representation for visual
  multi-agent reinforcement learning.
\newblock In {\em IROS}, 2021.

\bibitem{LiuNEURIPS2020}
Iou-Jen Liu, Raymond Yeh, and Alexander Schwing.
\newblock {High-Throughput Synchronous Deep RL}.
\newblock In {\em NeurIPS}, 2020.

\bibitem{LiuCORL2019}
Iou-Jen Liu, Raymond Yeh, and Alexander~G. Schwing.
\newblock {PIC: Permutation Invariant Critic for Multi-Agent Deep Reinforcement
  Learning}.
\newblock In {\em CoRL}, 2019.

\bibitem{sift}
D. Lowe.
\newblock Object recognition from local scale-invariant features.
\newblock {\em ICCV}, 1999.

\bibitem{Lowe2020}
Ryan Lowe, Abhinav Gupta, Jakob~N. Foerster, Douwe Kiela, and Joelle Pineau.
\newblock On the interaction between supervision and self-play in emergent
  communication.
\newblock In {\em ICLR}, 2020.

\bibitem{LoweNIPS2017}
R. Lowe, Y. Wu, A. Tamar, J. Harb, P. Abbeel, and I. Mordatch.
\newblock {Multi-Agent Actor-Critic for Mixed Cooperative-Competitive
  Environments}.
\newblock In {\em NeurIPS}, 2017.

\bibitem{MnihEtAlPMLR2016}
Volodymyr Mnih, Adria~Puigdomenech Badia, Mehdi Mirza, Alex Graves, Timothy
  Lillicrap, Tim Harley, David Silver, and Koray Kavukcuoglu.
\newblock Asynchronous methods for deep reinforcement learning.
\newblock In {\em ICML}, 2016.

\bibitem{MnihNature2015}
V. Mnih, K. Kavukcuoglu, D. Silver, A.~A. Rusu, J. Veness, M.~G. Bellemare, A.
  Graves, M. Riedmiller, A.~K. Fidjeland, G. Ostrovski, S. Petersen, C.
  Beattie, A. Sadik, I. Antonoglou, H. King, D. Kumaran, D. Wierstra, S. Legg,
  and D. Hassabis.
\newblock {Human-level control through deep reinforcement learning}.
\newblock {\em Nature}, 2015.

\bibitem{MordatchAAAI2018}
I. Mordatch and P. Abbeel.
\newblock {Emergence of Grounded Compositional Language in Multi-Agent
  Populations}.
\newblock In {\em AAAI}, 2018.

\bibitem{pyrobot2019}
Adithyavairavan Murali, Tao Chen, Kalyan~Vasudev Alwala, Dhiraj Gandhi, Lerrel
  Pinto, Saurabh Gupta, and Abhinav Gupta.
\newblock Pyrobot: An open-source robotics framework for research and
  benchmarking.
\newblock {\em arXiv preprint arXiv:1906.08236}, 2019.

\bibitem{Nair2018VisualRL}
Ashvin Nair, Vitchyr~H. Pong, Murtaza Dalal, Shikhar Bahl, S. Lin, and S.
  Levine.
\newblock Visual reinforcement learning with imagined goals.
\newblock In {\em NeurIPS}, 2018.

\bibitem{patel2021comon}
Shivansh Patel, Saim Wani, Unnat Jain, Alexander Schwing, Svetlana Lazebnik,
  Manolis Savva, and Angel Chang.
\newblock Interpretation of emergent communication in heterogeneous
  collaborative embodied agents.
\newblock {\em ICCV}, 2021.

\bibitem{pathakICMl17curiosity}
Deepak Pathak, Pulkit Agrawal, Alexei~A. Efros, and Trevor Darrell.
\newblock Curiosity-driven exploration by self-supervised prediction.
\newblock In {\em ICML}, 2017.

\bibitem{ramakrishnan2020occant}
Santhosh~K. Ramakrishnan, Ziad Al-Halah, and Kristen Grauman.
\newblock Occupancy anticipation for efficient exploration and navigation.
\newblock In {\em ECCV}, 2020.

\bibitem{ross2010efficient}
St{\'e}phane Ross and Drew Bagnell.
\newblock Efficient reductions for imitation learning.
\newblock In {\em AISTATS}, 2010.

\bibitem{RossAISTATS2011}
St{\'e}phane Ross, Geoffrey Gordon, and Drew Bagnell.
\newblock A reduction of imitation learning and structured prediction to
  no-regret online learning.
\newblock In {\em AISTATS}, 2011.

\bibitem{sammut1992learning}
Claude Sammut, Scott Hurst, Dana Kedzier, and Donald Michie.
\newblock Learning to fly.
\newblock In {\em Machine Learning Proceedings}, 1992.

\bibitem{habitat19iccv}
Manolis Savva, Abhishek Kadian, Oleksandr Maksymets, Yili Zhao, Erik Wijmans,
  Bhavana Jain, Julian Straub, Jia Liu, Vladlen Koltun, Jitendra Malik, Devi
  Parikh, and Dhruv Batra.
\newblock Habitat: {A} {P}latform for {E}mbodied {AI} {R}esearch.
\newblock In {\em ICCV}, 2019.

\bibitem{schulman2017proximal}
John Schulman, Filip Wolski, Prafulla Dhariwal, Alec Radford, and Oleg Klimov.
\newblock Proximal policy optimization algorithms.
\newblock {\em arXiv preprint arXiv:1707.06347}, 2017.

\bibitem{shridhar2020alfred}
Mohit Shridhar, Jesse Thomason, Daniel Gordon, Yonatan Bisk, Winson Han,
  Roozbeh Mottaghi, Luke Zettlemoyer, and Dieter Fox.
\newblock Alfred: A benchmark for interpreting grounded instructions for
  everyday tasks.
\newblock In {\em CVPR}, 2020.

\bibitem{Silver2018AGR}
D. Silver, T. Hubert, Julian Schrittwieser, Ioannis Antonoglou, Matthew Lai, A.
  Guez, Marc Lanctot, L. Sifre, D. Kumaran, T. Graepel, T. Lillicrap, K.
  Simonyan, and Demis Hassabis.
\newblock A general reinforcement learning algorithm that masters chess, shogi,
  and go through self-play.
\newblock {\em Science}, 2018.

\bibitem{Silver2017MasteringTG}
D. Silver, Julian Schrittwieser, K. Simonyan, Ioannis Antonoglou, Aja Huang, A.
  Guez, T. Hubert, L. Baker, Matthew Lai, A. Bolton, Yutian Chen, T. Lillicrap,
  F. Hui, L. Sifre, George van~den Driessche, T. Graepel, and Demis Hassabis.
\newblock Mastering the game of go without human knowledge.
\newblock {\em Nature}, 2017.

\bibitem{singh2020moca}
Kunal~Pratap Singh, Suvaansh Bhambri, Byeonghwi Kim, Roozbeh Mottaghi, and
  Jonghyun Choi.
\newblock Moca: A modular object-centric approach for interactive instruction
  following.
\newblock {\em arXiv preprint arXiv:2012.03208}, 2020.

\bibitem{szot2021habitat}
Andrew Szot, Alex Clegg, Eric Undersander, Erik Wijmans, Yili Zhao, John
  Turner, Noah Maestre, Mustafa Mukadam, Devendra Chaplot, Oleksandr Maksymets,
  et~al.
\newblock Habitat 2.0: Training home assistants to rearrange their habitat.
\newblock {\em arXiv preprint arXiv:2106.14405}, 2021.

\bibitem{thomason2020vision}
Jesse Thomason, Michael Murray, Maya Cakmak, and Luke Zettlemoyer.
\newblock Vision-and-dialog navigation.
\newblock In {\em CoRL}, 2020.

\bibitem{Trott2019KeepingYD}
A. Trott, Stephan Zheng, Caiming Xiong, and R. Socher.
\newblock Keeping your distance: Solving sparse reward tasks using
  self-balancing shaped rewards.
\newblock In {\em NeurIPS}, 2019.

\bibitem{wang2019reinforced}
Xin Wang, Qiuyuan Huang, Asli Celikyilmaz, Jianfeng Gao, Dinghan Shen,
  Yuan-Fang Wang, William~Yang Wang, and Lei Zhang.
\newblock Reinforced cross-modal matching and self-supervised imitation
  learning for vision-language navigation.
\newblock In {\em CVPR}, 2019.

\bibitem{wang2018look}
Xin Wang, Wenhan Xiong, Hongmin Wang, and William~Yang Wang.
\newblock Look before you leap: Bridging model-free and model-based
  reinforcement learning for planned-ahead vision-and-language navigation.
\newblock In {\em ECCV}, 2018.

\bibitem{wani2020multion}
Saim Wani, Shivansh Patel, Unnat Jain, Angel Chang, and Manolis Savva.
\newblock Multion: Benchmarking semantic map memory using multi-object
  navigation.
\newblock {\em NeurIPS}, 2020.

\bibitem{weihs2021visual}
Luca Weihs, Matt Deitke, Aniruddha Kembhavi, and Roozbeh Mottaghi.
\newblock Visual room rearrangement.
\newblock In {\em CVPR}, 2021.

\bibitem{WeihsJain2020Bridging}
Luca Weihs, Unnat Jain, Jordi Salvador, Svetlana Lazebnik, Aniruddha Kembhavi,
  and Alexander Schwing.
\newblock Bridging the imitation gap by adaptive insubordination.
\newblock {\em arXiv preprint arXiv:2007.12173}, 2020.

\bibitem{weihs2021learning}
Luca Weihs, Aniruddha Kembhavi, Kiana Ehsani, Sarah~M Pratt, Winson Han, Alvaro
  Herrasti, Eric Kolve, Dustin Schwenk, Roozbeh Mottaghi, and Ali Farhadi.
\newblock Learning generalizable visual representations via interactive
  gameplay.
\newblock In {\em ICLR}, 2021.

\bibitem{AllenAct}
Luca Weihs, Jordi Salvador, Klemen Kotar, Unnat Jain, Kuo-Hao Zeng, Roozbeh
  Mottaghi, and Aniruddha Kembhavi.
\newblock Allenact: A framework for embodied ai research.
\newblock {\em arXiv preprint arXiv:2008.12760}, 2020.

\bibitem{Wijmans2019EQAPhoto}
Erik Wijmans, Samyak Datta, Oleksandr Maksymets, Abhishek Das, Georgia
  Gkioxari, Stefan Lee, Irfan Essa, Devi Parikh, and Dhruv Batra.
\newblock {E}mbodied {Q}uestion {A}nswering in {P}hotorealistic {E}nvironments
  with {P}oint {C}loud {P}erception.
\newblock In {\em CVPR}, 2019.

\bibitem{wijmans2019dd}
Erik Wijmans, Abhishek Kadian, Ari Morcos, Stefan Lee, Irfan Essa, Devi Parikh,
  Manolis Savva, and Dhruv Batra.
\newblock Dd-ppo: Learning near-perfect pointgoal navigators from 2.5 billion
  frames.
\newblock In {\em ICLR}, 2019.

\bibitem{supp-williams1989learning}
Ronald~J Williams and David Zipser.
\newblock A learning algorithm for continually running fully recurrent neural
  networks.
\newblock {\em Neural computation}, 1989.

\bibitem{igibson}
Fei Xia, William~B. Shen, Chengshu Li, Priya Kasimbeg, Micael Tchapmi,
  Alexander Toshev, Roberto Martín-Martín, and Silvio Savarese.
\newblock Interactive gibson benchmark: A benchmark for interactive navigation
  in cluttered environments.
\newblock {\em IEEE Robotics and Automation Letters}, 2020.

\bibitem{xia2018gibson}
Fei Xia, Amir~R Zamir, Zhiyang He, Alexander Sax, Jitendra Malik, and Silvio
  Savarese.
\newblock Gibson env: Real-world perception for embodied agents.
\newblock In {\em CVPR}, 2018.

\bibitem{SAPIEN}
Fanbo Xiang, Yuzhe Qin, Kaichun Mo, Yikuan Xia, Hao Zhu, Fangchen Liu, Minghua
  Liu, Hanxiao Jiang, Yifu Yuan, He Wang, Li Yi, Angel~X. Chang, Leonidas~J.
  Guibas, and Hao Su.
\newblock Sapien: A simulated part-based interactive environment.
\newblock In {\em CVPR}, 2020.

\bibitem{ye2020auxiliary}
Joel Ye, Dhruv Batra, Erik Wijmans, and Abhishek Das.
\newblock Auxiliary tasks speed up learning pointgoal navigation.
\newblock In {\em CoRL}, 2020.

\bibitem{MetaWorld}
Tianhe Yu, Deirdre Quillen, Zhanpeng He, R. Julian, Karol Hausman, Chelsea
  Finn, and S. Levine.
\newblock Meta-world: A benchmark and evaluation for multi-task and meta
  reinforcement learning.
\newblock In {\em CoRL}, 2019.

\bibitem{zeng2021pushing}
Kuo-Hao Zeng, Luca Weihs, Ali Farhadi, and Roozbeh Mottaghi.
\newblock Pushing it out of the way: Interactive visual navigation.
\newblock In {\em CVPR}, 2021.

\bibitem{zhu2020vision}
Fengda Zhu, Yi Zhu, Xiaojun Chang, and Xiaodan Liang.
\newblock Vision-language navigation with self-supervised auxiliary reasoning
  tasks.
\newblock In {\em CVPR}, 2020.

\bibitem{ZhuARXIV2016}
Y. Zhu, R. Mottaghi, E. Kolve, J.~J. Lim, A. Gupta, L. Fei-Fei, and A. Farhadi.
\newblock {Target-driven Visual Navigation in Indoor Scenes using Deep
  Reinforcement Learning}.
\newblock In {\em ICRA}, 2017.

\end{thebibliography}
